\documentclass{article}

\PassOptionsToPackage{numbers, compress}{natbib}

\usepackage[final]{neurips_2022}

\usepackage[utf8]{inputenc} % allow utf-8 input
\usepackage[T1]{fontenc}    % use 8-bit T1 fonts
\usepackage{hyperref}       % hyperlinks
\usepackage{url}            % simple URL typesetting
\usepackage{booktabs}       % professional-quality tables
\usepackage{amsfonts}       % blackboard math symbols
\usepackage{nicefrac}       % compact symbols for 1/2, etc.
\usepackage{microtype}      % microtypography
\usepackage{xcolor}         % colors
\usepackage{amsmath}
\usepackage{amssymb}
\usepackage{graphicx}
\usepackage{wrapfig}
\usepackage{algorithm,algorithmic}
\usepackage{enumitem}

\newcommand{\vect}[1]{\mathbf{#1}}
\newcommand{\vectsymb}[1]{\boldsymbol{#1}}
\newcommand{\mat}[1]{\mathbf{#1}}
\newcommand{\normal}{\mathcal{N}}
\newcommand{\expect}{\mathbb{E}}
\newcommand{\norm}[1]{\lVert #1 \rVert}

\newcommand{\KL}{{\rm KL}}
\newcommand{\legendbox}[1]{\textcolor{#1}{\rule{2\fontcharht\font`X}{.8\fontcharht\font`X}}}

\definecolor{water}{rgb}{0,.77,1}
\definecolor{impervious}{rgb}{0.61,.61,0.61}
\definecolor{barren}{rgb}{1,.66,0}
\definecolor{trees}{rgb}{0.14,.44,0}
\definecolor{grass}{rgb}{0.63,1,0.45}

\title{Diffusion models as plug-and-play priors}
\author{%
    Alexandros Graikos \\
    Stony Brook University \\
    Stony Brook, NY \\
    \small\texttt{agraikos@cs.stonybrook.edu}
      \And
    Esmeralda S. Whitammer \\
    Mila, Universit\'e de Montr\'eal \\
    Montr\'eal, QC, Canada \\
    \small\texttt{esmeralda.whitammer@mila.quebec}
      \AND
    Nebojsa Jojic \\
    Microsoft Research \\
    Redmond, WA \\
    \small\texttt{jojic@microsoft.com}
      \And
    Dimitris Samaras \\
    Stony Brook University \\
    Stony Brook, NY \\
    \small\texttt{samaras@cs.stonybrook.edu}
}

\begin{document}

\maketitle

\begin{abstract}
    We consider the problem of inferring high-dimensional data $\vect{x}$ in a model that consists of a prior $p(\vect{x})$ and an auxiliary differentiable constraint $c(\vect{x},\vect{y})$ on $\vect{x}$ given some additional information $\vect{y}$. In this paper, the prior is an independently trained denoising diffusion generative model. The auxiliary constraint is expected to have a differentiable form, but can come from diverse sources. The possibility of such inference turns diffusion models into plug-and-play modules, thereby allowing a range of potential applications in adapting models to new domains and tasks, such as conditional generation or image segmentation.     The structure of diffusion models allows us to perform approximate inference by iterating differentiation through the fixed denoising network enriched with different amounts of noise at each step. Considering many noised versions of $\vect{x}$ in evaluation of its fitness is a novel search mechanism that may lead to new algorithms for solving combinatorial optimization problems. The code is available at \url{https://github.com/AlexGraikos/diffusion_priors}.
\end{abstract}

\section{Introduction}
\label{sec:introduction}
Deep generative models, such as denoising diffusion probabilistic models \cite[DDPMs;][]{sohl2015diffusion,ho2020ddpm} can capture the details of very complex distributions over high-dimensional continuous data $p(\vect{x})$ \cite{nichol2021improvedddpm,dhariwal2021diffusion,amit2021segdiff,sinha2021d2c,vahdat2021latent,hoogeboom2022equivariant}. The immense effective depth of DDPMs, sometimes with thousands of deep network evaluations in the generation process, is an apparent limitation on their use as off-the-shelf modules in hierarchical generative models, where models can be mixed and one model may serve as a prior for another conditional model. In this paper, we show that DDPMs trained on image data can be directly used as priors in systems that involve other differentiable constraints.

In our main problem setting, we assume that we have a prior $p(\vect{x})$ over high-dimensional data $\vect{x}$ and we wish to perform inference in a model that involves this prior and a constraint $c(\vect{x},\vect{y})$ on $\vect{x}$ given some additional information $\vect{y}$. That is, we want to find an approximation to the posterior distribution $p(\vect{x}|\vect{y}) \propto p(\vect{x})c(\vect{x},\vect{y})$. In this paper, $p(\vect{x}=\vect{x}_0,\vect{h}=\{\vect{x}_T, ..., \vect{x}_1\})$ is provided in the form of an independently trained DDPM over $\vect{x}_T,\dots,\vect{x}_0$ (\S\ref{sec:ddpm_prior}), making the DDPM a `plug-and-play' prior. 

Although the recent community interest in DDPMs has spurred progress in training algorithms and fast generation schedules \cite{nichol2021improvedddpm,salimans2022progressive,xiao2022tackling}, the possibility of their use as plug-and-play modules has not been explored. Furthermore, as opposed to existing work on plug-and-play models (starting from \cite{nguyen2017plugandplay}), the algorithms we propose {\em do not require} additional training or finetuning of model components or inference networks.

One obvious application of plug-and-play priors is conditional image generation (\S\ref{sec:mnist}, \S\ref{sec:faces}). For example, a denoising diffusion model trained on MNIST digit images might define $p(\vect{x})$, while the constraint $c(\vect{x},\vect{y})$ may be be the probability of digit class $\vect{y}$ under an off-the-shelf classifier. However, changing the semantics of $\vect{x}$, we can also use such models for inference tasks where neural networks struggle with domain adaptation, such as image segmentation: $c(\vect{x},\vect{y})$ constrains the segmentation $\vect{x}$ to match an appearance or a weak labeling $\vect{y}$ (\S\ref{sec:segmentation}). Finally, we describe a path towards using DDPM priors to solve continuous relaxations of combinatorial search problems by treating $\vect{y}$ as a latent variable with combinatorial structure that is deterministically encoded in $\vect{x}$ (\S\ref{sec:tsp}).

\subsection{Related work}
\label{sec:related_work}
\paragraph{Conditioning DDPMs.}
DDPMs have previously been used for conditional generation and image segmentation \cite{saharia2021superres,tashiro2021imputation,amit2021segdiff}. With few exceptions -- such as \cite{baranchuk2022labelefficient}, which uses a pretrained DDPM as a feature extractor -- these algorithms assume access to paired data and conditioning information during training of the DDPM model. In \cite{dhariwal2021diffusion}, a classifier $p(y \mid \vect{x}_t)$ that guides the denoising model towards the desired subset of images with the attribute $y$ is trained in parallel with the denoiser. In \cite{choi2021ilvr}, generation is conditioned on an auxiliary image by guiding the denoising process through correction steps that match the low-frequency components of the generated and conditioning images. In contrast, we aim to build models that combine an independently trained DDPM with an auxiliary constraint.

Our approach is also related to work on adversarial examples. Adversarial samples are produced by optimizing an image $\vect{x}$ to satisfy a desired constraint $c$ -- a classifier $p(\vect{y}|\vect{x})$ -- without reference to the prior over data. As supervised learning algorithms can ignore the structure in data $\vect{x}$, focusing only on the conditional distribution, it is possible to optimize for input $\vect{x}$ that provides the desired classification in various surprising ways \cite{adversarial_inputs}. In \cite{nie2022diffusion}, a diffusion model is used to defend from adversarial samples by making images more likely under a DDPM $p(\vect{x})$. We are instead interested in \emph{inference}, where we seek samples $\vect{x}$ that satisfy \emph{both} the classifier and the prior. (Our work may, however, have consequences for adversarial generation.) 

\paragraph{Conditional generation from unconditional models.} 
Works that preceded the recent popularity of DDPMs \cite{nguyen2017plugandplay,engel2018latent} show how an unconditional generative model, such as a generative adversarial network \cite[GAN;][]{goodfellow2014generative} or variational autoencoder \cite[VAE;][]{kingma2014auto}, can be combined with a constraint model to generate conditional samples. Regarding generative diffusion models, recent literature has focused on utilizing unconditional, pretrained DDPMs as priors to solve linear inverse imaging problems. Both in \cite{song2021solving} and \cite{kawar2021snips}, the authors modify the DDPM sampling algorithm, with knowledge of the linear degradation operator, to reconstruct an image consistent with the learned prior and given measurements. A generalization of these methods in \cite{kadkhodaie2021stochastic} shows how any pretrained denoising network can be used as the prior for solving linear inverse problems. We also clarify that although the term `plug-and-play' is widely used in the inverse imaging literature we refer to it in the scope of in-domain generation under differentiable constraints, in the same sense as \cite{nguyen2017plugandplay}.

\paragraph{Latent vectors in DDPMs.} Modeling the latent prior distribution in VAE-like models using a DDPM has been studied in \cite{sinha2021d2c, vahdat2021latent}. On the other hand, in \S\ref{sec:tsp}, we perform inference in the low-dimensional \emph{latent} space under a pretrained DDPM on a high-dimensional data space. Our approach to semantic segmentation (\S\ref{sec:segmentation}) is also related to \cite{rolf2022resolving}, where a prior $p(\vect{z})$ over latents is used to tune a posterior network $q(\vect{z}|\vect{x})$.  There, the priors are of relatively simple structure and are sample-specific, rather than global diffusion priors like in this paper.% Still, some practical problems can be tackled with both approaches .

\section{Method}
\label{sec:method}

\subsection{Problem setting}
\label{sec:problem_setting}
Recall that we want to find an approximation to the posterior distribution $p(\vect{x}|\vect{y}) \propto p(\vect{x})c(\vect{x},\vect{y})$, where $p(\vect{x})$ is a fixed prior distribution. Fixing $\vect{y}$ and introducing an approximate variational posterior $q(\vect{x})$, the free energy 
\begin{equation}
    F=-\expect_{q(\vect{x})} [\log p(\vect{x}) + \log c(\vect{x},\vect{y}) - \log q(\vect{x})]
    \label{eq:F_definition}
\end{equation}
is minimized when $q(\vect{x})$ is closest to the true posterior, i.e., when $\KL(q(\vect{x})\|p(\vect{x}|\vect{y}))$ is minimized. When $q(\vect{x})$, and the learning algorithm used to fit it, are expressive enough to capture the true posterior, this minimization yields the exact posterior $p(\vect{x}|\vect{y})$. Otherwise, $q$ will capture a `mode-seeking' approximation to the true posterior \cite{minka2005divergence}; in particular, if $q(\vect{y})$ is a Dirac delta, it is optimal to concentrate $q$ at the mode of $p(\vect{x}|\vect{y})$. When the prior involves latent variables $\vect{h}$ (i.e., $p(\vect{x})=\int_{\vect{h}}p(\vect{x}|\vect{h})p(\vect{h})\,d\vect{h}$), the free energy is 
\begin{align}
    F &=-\expect_{q(\vect{x})q(\vect{h}\mid \vect{x})} [\log p(\vect{x},\vect{h}) + \log c(\vect{x},\vect{y}) - \log q(\vect{x})q(\vect{h}|\vect{x})] \nonumber\\
    &=-\expect_{q(\vect{x})q(\vect{h}\mid \vect{x})} [\log p(\vect{x},\vect{h})  - \log q(\vect{x})q(\vect{h}|\vect{x})]-\expect_{q(\vect{x})} [\log c(\vect{x},\vect{y})] .
    \label{eq:F}
\end{align}
We are, in particular, interested in a general procedure for minimizing $F$ with respect to an approximate posterior $q(\vect{x})$ for any differentiable $c$ when $p$ is a DDPM (\S\ref{sec:ddpm_prior}).

A free energy of the same structure was also studied in \cite{vahdat2021latent}, where a DDPM $p(\vect{z})$ over a latent space is hybridized as a parent to a decoder $p(\vect{x}|\vect{z})$, with an additional inference model $q(\vect{z}|\vect{x})$  trained jointly with both of these models. On the other hand, we aim to work with independently trained components that operate directly in the pixel space, e.g., an off-the-shelf diffusion model $p(\vect{x})$ trained on images of faces and an off-the-shelf face classifier $p(\vect{y}|\vect{x})$, without training or finetuning them jointly (\S\ref{sec:faces}).

\subsection{Denoising diffusion probabilistic models as priors}
\label{sec:ddpm_prior}
Denoising diffusion probabilistic models (DDPMs) \cite{sohl2015diffusion,ho2020ddpm} generate samples $\vect{x}_0$ by reversing a (Gaussian) noising process. DDPMs are deep directed stochastic networks: 
\begin{align}
    p(\vect{x}_T, \vect{x}_{T-1},...,\vect{x}_0) &= p(\vect{x}_T)\prod_{t=1}^T p_{\theta}(\vect{x}_{t-1} \mid \vect{x}_t),
    \label{eq:P} \\
    p_{\theta}(\vect{x}_{t-1} \mid \vect{x}_t) &= \normal(\vect{x}_{t-1}; \vectsymb{\mu}_{\theta}(\vect{x}_t,t), \mat{\Sigma}_{\theta}(\vect{x}_t,t)),
    & p(\vect{x}_T) = \normal (\vect{0},\mat{I}),
    \label{eq:diff_reverse}
\end{align}
where $\vectsymb{\mu}_\theta$ and $\mat{\Sigma}_\theta$ are neural networks with learned parameters (often, as in this paper, $\mat{\Sigma}_\theta$ is fixed to a scalar diagonal matrix depending on $t$).
The model starts with a sample from a unit Gaussian $\vect{x}_T$ and successively transforms it with a nonlinear network $\vectsymb{\mu}_{\theta}(\vect{x}_t, t)$  adding a small Gaussian  innovation signal at each step according to a noise schedule.  After $T$ steps, the sample $\vect{x}=\vect{x}_0$ is obtained. 

In general, using such a model as a prior over $\vect{x}$ would require an intractable integration over latent variables $\vect{h}=(\vect{x}_T,...,\vect{x}_1)$:
\begin{equation}
    p(\vect{x})= \int_{\vect{h}} p(\vect{x}_T, \vect{x}_{T-1},...,\vect{x}_1, \vect{x}_0=\vect{x})\,d\vect{x}_T\,\dots\,d\vect{x}_1.
    \label{eq:marginal}
\end{equation}
However, DDPMs are trained under the assumption that the posterior $q(\vect{x}_t|\vect{x}_{t-1})$ is a simple diffusion process that successively adds Gaussian noise according to a predefined schedule $\beta_t$:
\begin{equation}
    q(\vect{x}_t \mid \vect{x}_{t-1}) = \normal(\vect{x}_t ; \sqrt{1-\beta_t}\vect{x}_{t-1}, \beta_t \mat{I}),\quad t=1,\ldots,T.
    \label{eq:diff_forward}
\end{equation}
Therefore, if $p(\vect{x})$ is the likelihood (\ref{eq:marginal}) of $\vect{x}$ under a DDPM, then in the first expectation of (\ref{eq:F}) we should use $q(\vect{h}=\{\vect{x}_T,...,\vect{x}_1\}|\vect{x}_0=\vect{x})=\prod_{t=1}^T  q(\vect{x}_t \mid \vect{x}_{t-1})$. The simplest approximation to the posterior over $\vect{x}=\vect{x}_0$ is a point estimate:
\begin{equation}
    q(\vect{x})= \delta (\vect{x} - \vectsymb{\eta})
    \label{eq:q_def}
\end{equation}
where by $\delta$ we denote the Dirac delta function. Thus, we can sample $\vect{x}_t$ at any arbitrary time step using the forward noising process as
\begin{equation}
    q(\vect{x}_t) = \normal(\vect{x}_t ; \sqrt{\bar{\alpha}_t}\vectsymb{\eta}, (1-\bar{\alpha}_t) \mat{I})
    \label{eq:diff_sample}
\end{equation}
where $\alpha_t = 1 - \beta_t$ and $\bar{\alpha}_t = \prod_{i=1}^t \alpha_t$. Analogously to \cite{ho2020ddpm}, we can also extract a conditional Gaussian  $q(\vect{x}_{t-1} \mid \vect{x}_t, \vectsymb{\eta})$ and express the first expectation in (\ref{eq:F}) as
\begin{equation}
    -\expect_{q(\vect{x})q(\vect{h}\mid \vect{x})} [\log p(\vect{x},\vect{h})  - \log q(\vect{x})q(\vect{h}|\vect{x})] = \sum_t \KL(q(\vect{x}_{t-1} \mid \vect{x}_t, \vectsymb{\eta}) \,\|\, p_{\theta}(\vect{x}_{t-1} \mid \vect{x}_t)),
    \label{eq:sum_of_KL}
\end{equation}
which after reparametrization \cite{ho2020ddpm} leads to
\begin{equation}
    \sum_t w_t(\beta)\expect_{\vectsymb{\epsilon} \sim \normal (\vect{0},\mat{I})}
    [\norm{\vectsymb{\epsilon} - \vectsymb{\epsilon}_{\theta}(\vect{x}_t,t)}_2^2],\quad 
    \vect{x}_t = \sqrt{\bar{\alpha}_t}\vectsymb{\eta} + \sqrt{1-\bar{\alpha}_t}\vectsymb{\epsilon},
    \label{eq:reparametrization}
\end{equation}
where the stage $t$ noise reconstruction $\vectsymb{\epsilon}_{\theta}(\vect{x}_t,t)$ is a linear transformation of the model's expectation $\vectsymb{\mu}_{\theta}(\vect{x}_t,t)$: 
\begin{equation}
    \vectsymb{\mu}_\theta(\vect{x}_t,t)=\frac{1}{\sqrt{\alpha_t}}\left(\vect{x}_t-\frac{\beta_t}{\sqrt{1-\bar\alpha_t}}\vectsymb{\epsilon}_\theta(\vect{x}_t,t)\right).
    \label{eq:reparametrization_mu}
\end{equation}
The weighting $w_t(\beta)$ is generally a function of the noise schedule, but in most pretrained diffusion models it is set to 1. Thus, the free energy in (\ref{eq:F}) reduces to
\begin{align}
    F&= \sum_t \expect_{\vectsymb{\epsilon} \sim \normal (\vect{0},\mat{I})}
    [\norm{\vectsymb{\epsilon} - \vectsymb{\epsilon}_{\theta}(\vect{x}_t,t)}_2^2]-\expect_{q(\vect{x})} [\log c(\vect{x},\vect{y})]
    \nonumber\\
    &= \sum_t \expect_{\vectsymb{\epsilon} \sim \normal (\vect{0},\mat{I})}
    [\norm{\vectsymb{\epsilon} - \vectsymb{\epsilon}_{\theta}(\vect{x}_t,t)}_2^2]- \log c(\vectsymb{\eta},\vect{y}), \quad 
    \vect{x}_t = \sqrt{\bar{\alpha}_t}\vectsymb{\eta} + \sqrt{1-\bar{\alpha}_t}\vectsymb{\epsilon}.
    \label{eq:dirac_opt}
\end{align}
The first term is the cost usually used to learn the parameters $\theta$ of the diffusion model. To perform inference under an already trained model $\vectsymb{\epsilon}_\theta$, \emph{we instead minimize $F$ with respect to $\vectsymb{\eta}$ through sampling $\vectsymb{\epsilon}$ in the summands over $t$}.

A similar derivation applies if a Gaussian approximation to the posterior $q(\vect{x})$ is used (see \S\ref{sec:gaussian_estimate}). Such an approximation allows to model not only a mode of the posterior, but the uncertainty in its vicinity.

We summarize the algorithm for a point estimate $q(\vect{x})$ as Algorithm~\ref{alg:the_algorithm}. Variations on this algorithm are possible. Depending on how close to a good mode we can initialize $\vectsymb{\eta}$, this optimization may involve summing only over $t\le t_{\rm max}<T$; different time step schedules can be considered depending on the desired diversity in the estimated $\vect{x}$. Note that optimization is stochastic and each time it is run it can produce different point estimates of $\vect{x}$ which are are both likely under the diffusion prior and satisfy the constraint as much as possible. 

We observed that optimizing simultaneously for all $t$ makes it difficult to guide the sample towards a mode in image generation applications; therefore, we anneal $t$ from high to low values. Intuitively, the first few iterations of gradient descent should coarsely explore the search space, while later iterations gradually reduce the temperature to steadily reach a nearby local maximum of $p(\vect{x}|\vect{y})$. Examples of annealing schedules designed for the tasks demonstrated in \S\ref{sec:experiments_image}, \ref{sec:segmentation}, \ref{sec:tsp} are presented in the Appendix (Fig.~\ref{fig:annealing_schedules}).

Another interesting case is when  $\vect{x}$ is parametrized through a latent variable (this can be seen as a case of a hard, non-differentiable constraint: if $\vect{x}$ is a deterministic function of $\vect{y}$, $\vect{x}=f(\vect{y})$, then $c(\vect{x},\vect{y})$ is supported on the corresponding manifold). Then the procedure in Algorithm~\ref{alg:the_algorithm} can be performed with gradient descent steps with respect to $\vect{y}$ on 
\begin{equation}
    \norm{\vectsymb{\epsilon} - \vectsymb{\epsilon}_{\theta}(\sqrt{\bar{\alpha}_{t_i}}f(\vect{y})+\sqrt{1-\bar{\alpha}_{t_i}}\vectsymb{\epsilon},t_i)}_2^2
    \label{eq:latent_objective}
\end{equation} instead of steps 4 and 5. (For some semantics of the latent representation, one may wish to make the prior on $\vect{x}$ the pushforward by $f$ of a known prior on the latent $\vect{y}$. In this case, (\ref{eq:latent_objective}) must be weighted by the Jacobian of $f$ at $\vect{y}$.)

\begin{algorithm}[t]
    \begin{algorithmic}[1]
        \INPUT pretrained DDPM $\vectsymb{\epsilon}_\theta$, auxiliary data $\vect y$, constraint $c$, time schedule $(t_i)_{i=1}^T$, learning rate $\lambda$
        \STATE Initialize $\vect{x}\sim\cal N(\vectsymb{0};\mat{I})$.
        \FOR{$i=T..1$}
        \STATE Sample $\vectsymb{\epsilon}\sim\mathcal{N}(\vectsymb{0};\mat{I})$
        \STATE $\vect{x}_{t_i}=\sqrt{\bar{\alpha}_{t_i}}\vect{x}+\sqrt{1-\bar{\alpha}_{t_i}}\vectsymb{\epsilon}$
        \STATE $\vect{x} \leftarrow  \vect{x} 
            - \lambda\nabla_{\vect{x}}
            [\norm{\vectsymb{\epsilon} - \vectsymb{\epsilon}_{\theta}(\vect{x}_{t_i},t_i)}_2^2  -\log c(\vect{x},\vect{y})]$
        \ENDFOR 
        \OUTPUT $\vectsymb{\eta}=\vect{x}$
    \end{algorithmic}
    \caption{Inferring a point estimate of $p(\vect{x}|\vect{y})\approx\delta(\vect{x} - \vectsymb{\eta})$, under a DDPM prior and constraint.}
    \label{alg:the_algorithm}
\end{algorithm}

\section{Experiments: Conditional image generation}
\label{sec:experiments_image}

\subsection{Simple illustration on MNIST}
\label{sec:mnist}

\begin{figure}[t]
    \centering
    \small
    \begin{tabular} {cccccc}
        \includegraphics[width=0.1\textwidth]{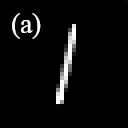} &
        \includegraphics[width=0.1\textwidth]{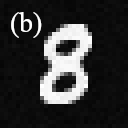} &
        \includegraphics[width=0.1\textwidth]{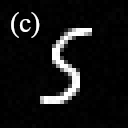} &
        \includegraphics[width=0.1\textwidth]{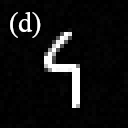} &
        \includegraphics[width=0.1\textwidth]{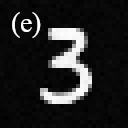} &
        \includegraphics[width=0.1\textwidth]{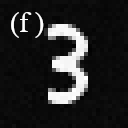} \\
        Thin & Thick & Vert. & Horiz. & Class `3' & Class `3' \& \\
         & & Asymmetry & Asymmetry & & Symmetry \\
    \end{tabular}
    \caption{Inferred MNIST samples under different conditions $c(\vect{x},\vect{y})$.}
    \label{fig:mnist_samples}
\end{figure}

We first explore the idea of generating conditional samples from an unconditional diffusion model on MNIST. We train the DDPM model of \cite{dhariwal2021diffusion} on MNIST digits and experiment with different sets of constraints $\log c(\vect{x},\vect{y})$ to generate samples with specific attributes. The examples in Fig. \ref{fig:mnist_samples} showcase such generated samples. For the digit in (a) we set the constraint $\log c$ to be the unnormalized score of `thin' digits, computed as negative of the average image intensity, whereas in (b) we invert that and generate a `thick' digit with high mean intensity. Similarly, in (c) and (d) we hand-craft a score that penalizes the vertical and horizontal symmetry respectively, by computing the $L^2$ distance between the two folds (vertical/horizontal) of the digit $\vect{x}$, which leads to the generation of skewed, non-symmetric samples.

We also showcase how the auxiliary constraint $c(\vect{x},\vect{y})$ can be modeled by a different, independently trained network. The digit in Fig.~\ref{fig:mnist_samples} (e) is generated by constraining the DDPM with a classifier network that is separately trained to distinguish between the digit class $\vect{y}=3$ and all other digits. The auxiliary constraint in this case is the likelihood of the inferred digit, as it is estimated by the classifier. Finally, for (f) we multiply horizontal \textit{symmetry} and digit classifier constraints, prompting the inference procedure to generate a perfectly centered and symmetric digit. Details of model training and inference can be found in the Appendix (\S\ref{sec:mnist_appendix}).

\subsection{Using off-the-shelf components for conditional generation of faces}
\label{sec:faces}

We consider the generation of natural images with a pretrained DDPM prior and a learned constraint. We utilize the pretrained DDPM network on FFHQ-256 \cite{karras2019stylegan} from \cite{baranchuk2022labelefficient} and a pretrained ResNet-18 face attribute classifier on CelebA \cite{liu2015faceattributes}. The attribute classifier computes the likelihood of presence of various facial features $y$ in a given image $\vect{x}$, as they are defined by the CelebA dataset. Examples of such features are \textit{no beard}, \textit{smiling}, \textit{blond hair} and \textit{male}. To generate a conditional sample from the unconditional DDPM network we select a subset of these and enforce their presence or absence using the classifier predicted likelihoods as our constraint $c$. If $\vect{y}$ is a set of attributes we wish to be present, the constraint $\log c(\vect{x},\vect{y})$ can be expressed as
\begin{equation}
    \log c(\vect{x},\vect{y}) = \sum_{y \in \vect{y}} \log p(y\mid \vect{x})
    \label{eq:face_constraint}
\end{equation}
We only strictly enforce a small subset of facial attributes and therefore $\vect{x}$ is allowed to converge towards different modes that correspond to samples that exhibit, in varying levels, the desired features.

\begin{figure}[t]
    \centering
    \small
    \begin{tabular}{@{}c@{\hspace{1mm}}@{\hspace{1mm}}c@{\hspace{2mm}}c@{\hspace{2mm}}c@{\hspace{2mm}}c@{\hspace{2mm}}c@{}}
        \includegraphics[width=0.15\textwidth]{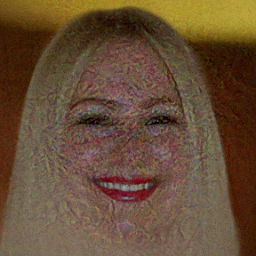} &
        \includegraphics[width=0.15\textwidth]{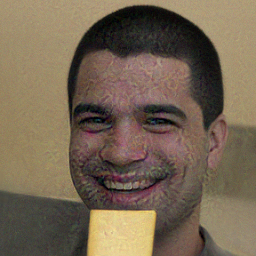} &
        \includegraphics[width=0.15\textwidth]{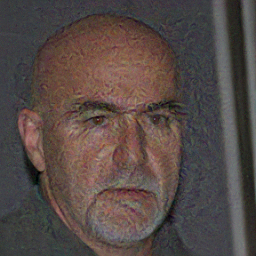} &
        \includegraphics[width=0.15\textwidth]{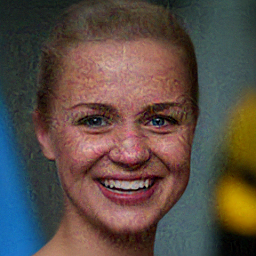} &
        \includegraphics[width=0.15\textwidth]{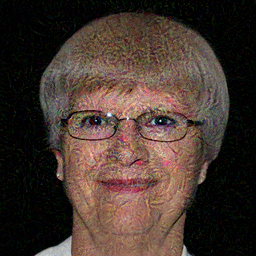} &
        \includegraphics[width=0.15\textwidth]{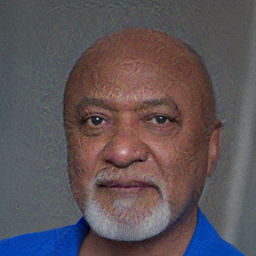} \\
        \includegraphics[width=0.15\textwidth]{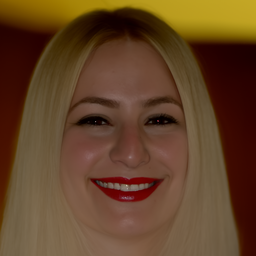} &
        \includegraphics[width=0.15\textwidth]{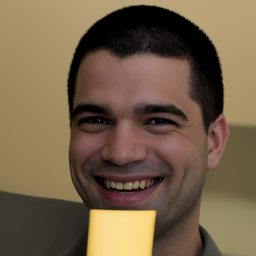} &
        \includegraphics[width=0.15\textwidth]{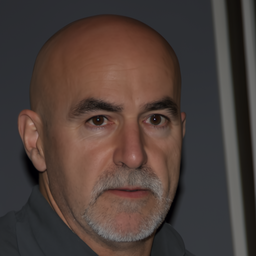} & 
        \includegraphics[width=0.15\textwidth]{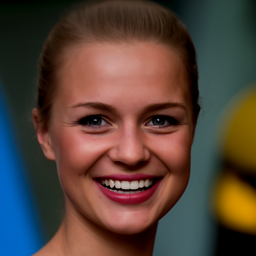} &
        \includegraphics[width=0.15\textwidth]{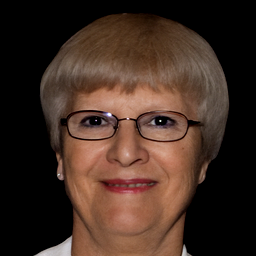} &
        \includegraphics[width=0.15\textwidth]{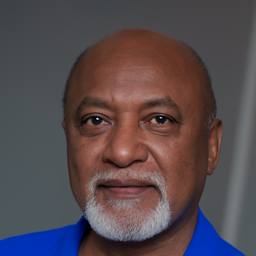} \\
        Blonde & Five-o'clock & Oval & High & Eyeglasses & Goatee \& \\
        & Shadow & & Cheekbones & & Big Nose
    \end{tabular}
    \caption{First row: Conditional FFHQ samples $\vect{x}$ for constraints $c(\vect{x},\vect{y})$ with various attribute sets $\vect{y}$. Second row: denoising as in \cite{nie2022diffusion} to remove artifacts that appear when optimizing with a classifier network enforcing the constraint. }
    \label{fig:face_samples}
\end{figure}

In Fig.~\ref{fig:face_samples} we demonstrate our ability to infer conditional samples $\vect{x}$ with desired attributes $\vect{y}$, using only the unconditional diffusion model and the classifier $p(\vect{y} \mid \vect{x})$. In the first row, we show the results of the optimization procedure of Algorithm \ref{alg:the_algorithm} for various attributes. The classifier objective $c(\vect{x},\vect{y})$ manipulates the image with the goal of making the classifier network produce the desired attribute predictions, whereas the diffusion objective attempts to pull the sample $x$ towards the learned distribution $p(\vect{x})$. If we ignored the denoising loss, the result would be some adversarial noise that fools the classifier network. The DDPM prior, however, is strong enough to guide the process towards realistic-looking images that simultaneously satisfy the classifier constraint set. 

We notice that the generated samples $\vect{x}$, although having converged towards a correct mode of $p(\vect{x})$, still exhibit a noticeable amount of noise related to the optimization of classifier objective. To address that, inspired by \cite{nie2022diffusion}, we simply denoise the image using the DDPM model alone, starting from the low noise level $t=200$ so as to retain the overall structure. The results of this denoising are shown in the second row of Fig.~\ref{fig:face_samples}. 

\begin{figure}[t]
    \centering
    \small
    \begin{tabular}{c@{\hspace{1mm}}c@{\hspace{1mm}}c@{\hspace{1mm}}c@{\hspace{1mm}}c@{\hspace{1mm}}c@{\hspace{1mm}}c@{\hspace{1mm}}c@{\hspace{1mm}}c@{\hspace{1mm}}c@{\hspace{1mm}}}
        $t=1000$ & $t=962$ & $t=896$ & $t=807$ & $t=701$ & $t=585$ & $t=465$ & $t=349$ & $t=242$ & Denoise \\
        \includegraphics[width=0.09\textwidth]{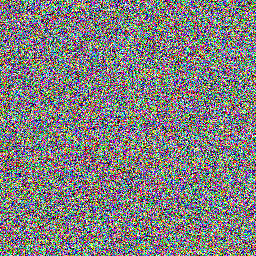} &
        \includegraphics[width=0.09\textwidth]{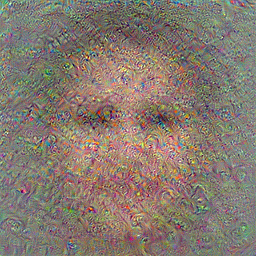} &
        \includegraphics[width=0.09\textwidth]{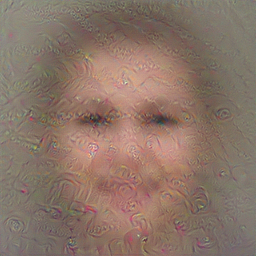} &
        \includegraphics[width=0.09\textwidth]{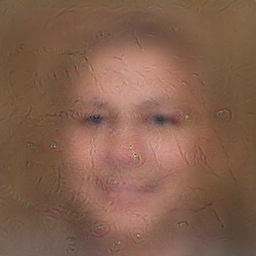} &
        \includegraphics[width=0.09\textwidth]{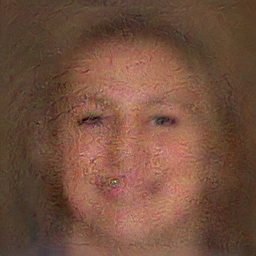} &
        \includegraphics[width=0.09\textwidth]{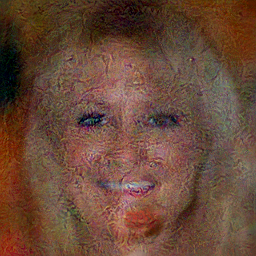} &
        \includegraphics[width=0.09\textwidth]{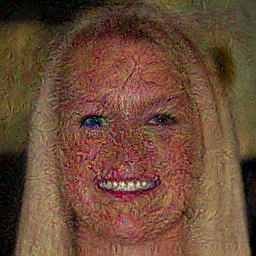} &
        \includegraphics[width=0.09\textwidth]{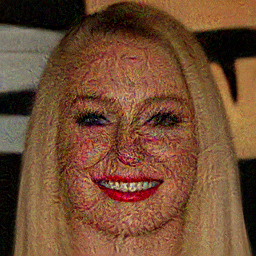} &
        \includegraphics[width=0.09\textwidth]{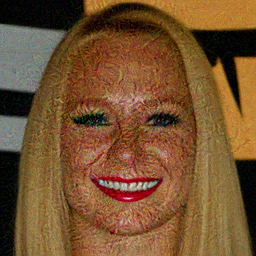} &
        \includegraphics[width=0.09\textwidth]{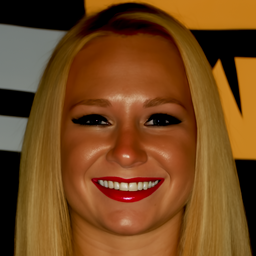} \\
    \end{tabular}
    \caption{FFHQ conditional generation for $\vect{y}=\{$\textit{Blonde},\ \textit{Smiling},\ \textit{Female}$\}$. The last step performs denoising as in \cite{nie2022diffusion} to remove artifacts that appear when training on a classifier as a constraint.}
    \label{fig:face_steps}
\end{figure}

In Fig.~\ref{fig:face_steps} we showcase the intermediate steps of the optimization process for inference with the conditions \textit{blond hair}+\textit{smiling}+\textit{not male}, thus solving a problem like that studied in \cite{du2020compositional} using only \emph{independently trained} attribute classifiers and an unconditional generative model of faces. The sample $x$ is initialized with Gaussian noise $\normal(\vect{0},\mat{I})$, and as we perform gradient steps with decreasing values of $t$, we observe facial features being added in a coarse-to-fine manner.

In the Appendix (\S\ref{sec:ffhq_appendix}) we provide additional samples and further discuss the sample quality in comparison to unconditional generation. We also present results on inference with conflicting attributes as well as common failure cases.

\section{Experiments: Semantic image segmentation}
\label{sec:segmentation}
We test the applicability of diffusion priors in discrete tasks, such as inferring semantic segmentations from images. For this purpose, we use the EnviroAtlas dataset \cite{pickard2015enviroatlas} which is composed of 5-class, 1m-resolution land cover labels from four geographically diverse cities across the US; Pittsburgh, PA, Durham, NC, Austin, TX and Phoenix, AZ. We only have access to the high resolution labels from Pittsburgh, and the task is to infer the land cover labels in the other three cities, given only probabilistic weak labels $\ell_{\rm weak}$ derived from coarse auxiliary data \cite{rolf2022resolving}. We use Algorithm~\ref{alg:the_algorithm} to perform an inference procedure that does not directly take imagery as input, but uses constraints derived from unsupervised color clustering. We use only cluster indices in inference, making the algorithm dependent on image structure, but not color. Local cluster indices as a representation have a promise of extreme domain transferability, but they require a form of a combinatorial search which matches local cluster indices to semantic labels so that the created shapes resemble previously observed land cover, as captured by a denoising diffusion model of semantic segmentations.  

\begin{figure}[t]
    \centering
    \begin{tabular}{c c c c c}
        \includegraphics[width=0.15\textwidth]{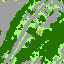} &
        \includegraphics[width=0.15\textwidth]{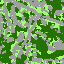} &
        \includegraphics[width=0.15\textwidth]{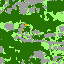} &
        \includegraphics[width=0.15\textwidth]{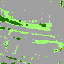} &
        \includegraphics[width=0.15\textwidth]{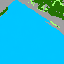} \\
    \end{tabular} \\
    {\footnotesize
    \legendbox{water} Water
    \legendbox{impervious} Impervious Surface
    \legendbox{barren} Soil and Barren
    \legendbox{trees} Trees and Forest
    \legendbox{grass} Grass and Herbaceous}
    \caption{Unconditional samples from the DDPM trained on land cover segmentations (cf.~Fig.~\ref{fig:seg_inference_samples}).}
    \label{fig:seg_samples}
\end{figure}

\paragraph{DDPM on semantic pixel labels.} We train a DDPM model on the $\frac{1}{4}$-resolution one-hot representations of the land cover labels, using the U-Net diffusion model architecture from \cite{dhariwal2021diffusion}. To convert the one-hot diffusion samples to probabilities we follow \cite{hoogeboom2022equivariant} and assume that for any pixel $i$ in the inferred sample $\vect{x}$, the distribution over the label $\ell$ is, $p(\ell_i) \propto \int_{0.5}^{1.5} \normal(x_{i}^{\ell} \mid \eta_i, \sigma)$, where $\sigma$ is user-defined a parameter. We chose this approach for its simplicity and ease to apply in our inference setting of Algorithm \ref{alg:the_algorithm}. Alternatively, we could use diffusion models for categorical data \cite{hoogeboom2021argmax} with the appropriate modifications to our inference procedure. Samples drawn from the learned distribution are presented in Fig \ref{fig:seg_samples}. 

\begin{figure}[t]
    \centering
    \small
    \begin{tabular}{ccccc}
        & & Weak & & \\
        Image & Clustering $\vect{z}$ & Labels $\ell_{\rm weak}$ & Inferred $\vect{x}$ & Ground Truth \\
        \includegraphics[width=0.15\textwidth]{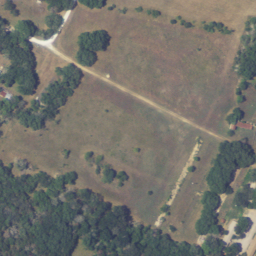} &
        \includegraphics[width=0.15\textwidth]{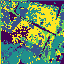} &
        \includegraphics[width=0.15\textwidth]{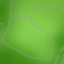} &\includegraphics[width=0.15\textwidth]{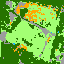}
         &
        \includegraphics[width=0.15\textwidth]{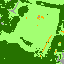} \\
        \includegraphics[width=0.15\textwidth]{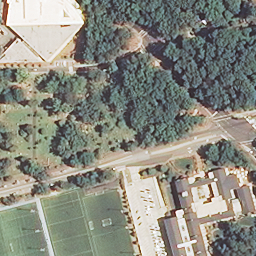} &
        \includegraphics[width=0.15\textwidth]{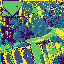} &
        \includegraphics[width=0.15\textwidth]{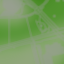} &
        \includegraphics[width=0.15\textwidth]{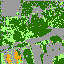} &\includegraphics[width=0.15\textwidth]{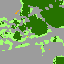}
         \\
        \includegraphics[width=0.15\textwidth]{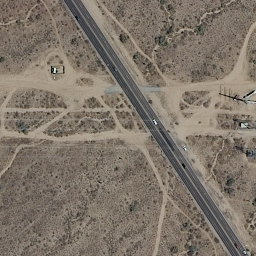} &
        \includegraphics[width=0.15\textwidth]{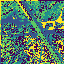} &
        \includegraphics[width=0.15\textwidth]{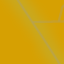} &
        \includegraphics[width=0.15\textwidth]{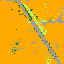} & \includegraphics[width=0.15\textwidth]{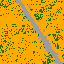} \\
    \end{tabular}
    \caption{Segmentation inference results. The inferred segmentation $\vect{x}$ is initialized with the weak labels to reduce the number of steps needed. The samples are chosen from (top to bottom) Durham, NC, Austin, TX and Phoenix, AZ. Although AZ has a vastly different joint distribution of colors and labels, the inferred segmentation still captures the overall structure. Note that the inference algorithm does not use the pixel intensities in the input image, only an unsupervised color clustering.}
    \label{fig:seg_inference_samples}
\end{figure}

\paragraph{Inferring semantic segmentations.} In order to infer the segmentation of a single image, under the diffusion prior, we directly apply Algorithm \ref{alg:the_algorithm} with a hand-crafted constraint $c$ which provides structural and label guidance. To construct $c$, we first compute a local color clustering $\vect{z}$ of input the image (\S\ref{sec:land_cover_appendix} in the Appendix). In addition, we utilize the available weak labels $\vect{\ell}_{\rm weak}$ \cite{rolf2022resolving} and force the predicted segments' distribution to match the weak label distribution when averaged in non-overlapping blocks. We combine the two objectives in a single constraint $c(\vect{x},\vect{z},\vect{\ell}_{\rm weak})$ by (i) computing the mutual information between the color clustering $\vect{z}$ and the predicted labels $\vect{x}$ , transformed into a valid probability distribution from the inferred one-hot vectors, in overlapping image patches and (ii) computing the negative KL divergence between the average predicted distribution and the distribution given by the weak labels in non-overlapping blocks 
\begin{equation}
    \log c(\vect{x},\vect{z},\ell_{\rm weak}) = {\rm MI}(\vect{x},\vect{z}) - {\rm KL}(\vect{x} \,\|\,\ell_{\rm weak}).
    \label{eq:segmentation_constraints}
\end{equation}
Empirically, we find that we can reduce the number of optimization steps needed to perform inference by initializing the sample $\vect{x}$ with the weak labels $\ell_{\rm weak}$ instead of random noise, allowing us to start from a smaller $t_i$. Examples of images and their inferred segmentations are shown in Fig.~\ref{fig:seg_inference_samples}.

\paragraph{Domain transfer with inferred samples.} The above inference procedure is agnostic to colors by design, and we expect it to have a greater ability to perform in new areas than the approach in \cite{rolf2022resolving}, which still finetunes networks that take raw images as input. We also investigate domain transfer approaches where patches segmented using the the diffusion prior are used to train neural networks for fast inference. We pretrain a standard U-Net inference network $p(\vect{x} \mid I)$ solely on 20k batches of 16 randomly sampled $64\times64$ image patches in PA. We randomly sample 640 images in each of the other geographies and generate semantic segmentations using our inference procedure, then finetune the inference network on these segmentations. This network is then evaluated on the entire target geography.

The results  in Table \ref{tab:enviroatlas} demonstrate that this approach to domain transfer is comparable with the state-of-the-art work of \cite{rolf2022resolving} for weakly-supervised training. The na\"ive approach of training a U-Net only on the available high-resolution PA data (PA supervised) fails to generalize to the geographically different location of Phoenix, AZ. Similarly, the model of \cite{robinson2019large}, which is a US-wide high-resolution land cover model trained on imagery and labels, and multi-resolution auxiliary data over the entire contiguous US also suffers. When the weak labels are provided as input (PA supervised + weak) the results can improve significantly.

\begin{table}[t]
    \caption{Accuracies and class mean intersection-over-union scores on the EnviroAtlas dataset in various geographic domains. The model in the second-to-last row was pretrained in a supervised way on labels in the Pittsburgh, PA, region.}
    \label{tab:enviroatlas}
    \centering
    \begin{tabular}{lcccccc}
\toprule
&\multicolumn{2}{c}{Durham, NC}&\multicolumn{2}{c}{Austin, TX}&\multicolumn{2}{c}{Phoenix, AZ}\\
\cmidrule(lr){2-3}\cmidrule(lr){4-5}\cmidrule(lr){6-7}
Algorithm&Acc \%&IoU \%&Acc \%&IoU \%&Acc \%&IoU \%\\\midrule
% these are the QR (r) predictions
PA supervised & 74.2 & 35.9 & 71.9 & 36.8 & 6.7 & 13.4 \\
PA supervised + weak & 78.9 & 47.9 & 77.2 & 50.5 & 62.8 & 24.2  \\
Implicit posterior \cite{rolf2022resolving} & 79.0 & 48.4  & 76.6  & 49.5 & 76.2 & 46.0 \\
\midrule
Ours (from scratch) & 76.0 & 39.9 & 74.8 & 39.4 & 69.5 & 31.6 \\
Ours (fine-tuned) & 79.8 & 46.4 & 79.5 & 45.4 & 69.6 & 32.4 \\\midrule
Full US supervised \cite{robinson2019large} & 77.0 & 49.6 & 76.5 & 51.8 & 24.7 & 23.6 \\
\bottomrule
\end{tabular}

\end{table}

\section{Experiments: Continuous relaxation of combinatorial problems}
\label{sec:tsp}

So far, we have considered inference under a DDPM prior and a differentiable constraint $c(\vect{x},\vect{y})$. We consider the case of a `hard' constraint, where $\vect{y}$ is a latent vector deterministically encoded in an image $\vect{x}$ ($\vect{x}=f(\vect{y})$) and we have a DDPM prior over images $p_{\rm DDPM}(\vect{x})$. We will use the variation of Algorithm~\ref{alg:the_algorithm} described at the end of \S\ref{sec:ddpm_prior} to obtain a point estimate of the distribution over $y$, $p(\vect{y})\propto p_{\rm DDPM}(f(\vect{y}))$.

We illustrate this in the setting of a well-known combinatorial problem, the traveling salesman problem (TSP). Recall that a Euclidean traveling salesman problem on the plane is described by $N$ points $v_1,\dots,v_N\in\mathbb{R}^2$, which form the vertex set of a complete weighted graph $G$, where the weight of the edge from $v_i$ to $v_j$ is the Euclidean distance $\norm{v_i-v_j}$. A \emph{tour} of $G$ is a connected subgraph in which every vertex has degree 2. The TSP is the optimization problem of finding the tour with minimal total weight of the edges, or, equivalently, a permutation $\sigma$ of $\{1,2,\dots,N\}$ that minimizes
\[
\norm{v_{\sigma(1)}-v_{\sigma(2)}}
+\norm{v_{\sigma(2)}-v_{\sigma(3)}}
+\dots
+\norm{v_{\sigma(N-1)}-v_{\sigma(N)}}
+\norm{v_{\sigma(N)}-v_{\sigma(1)}}.
\]
Although the general form of the TSP is NP-hard, a polynomial-time approximation scheme is known to exist in the Euclidean case \cite{arora1998polynomial,mitchell1999guillotine} and can yield proofs of tour optimality for small problems.

Humans have been shown to have a natural propensity for solving the Euclidean TSP (see \cite{macgregor2011human} for a survey). Humans construct a tour by processing an image representation of the points $v_1,\dots,v_N$ through their visual system. However, the optimization algorithms in common use for solving the TSP do not use a vision inductive bias, instead falling into two broad categories:
\begin{itemize}[leftmargin=*,itemsep=0pt,topsep=0pt]
    \item Discrete combinatorial optimization algorithms and efficient integer programming solvers, studied for decades in the optimization literature \cite{lkh,helsgaun,concorde};
    \item More recently, there has been work on neural nets, trained by reinforcement learning or imitation learning, that build tours sequentially or learn heuristics for their (discrete) iterative refinement. Successful recent approaches \cite{deudon2018learning,kool2019attention,joshi2019efficient,joshi2021learning,bresson2021tsptransformer} have used Transformer \cite{transformer} and graph neural network \cite{gnn} architectures.
\end{itemize}
The algorithm we propose using DDPMs is a hybrid of these categories: it reasons over a continuous relaxation of the problem, but exploits the learning of generalizable structure in example solutions by a neural model. In addition, ours is the first TSP algorithm to mimic the convolutional inductive bias of the visual system. 

\paragraph{Encoding function.}
Fix a set of points $v_1,\dots,v_N\in[0,1]\times[0,1]$. We encode an symmetric $N\times N$ matrix with 0 diagonal $A$ as a $64\times64$ greyscale image $f(A)$ by superimposing: (i) raster images of line segments from $v_i$ to $v_j$ with intensity value $A_{ij}$ for every pair $(i,j)$, and (ii) raster images of small black dots placed at $v_i$ for each $i$. For example, if $A$ is the adjacency matrix of a tour, then $f(A)$ is a visualization of this tour as a $64\times64$ image.

\begin{wrapfigure}[13]{r}{0.5\textwidth}
    \centering
    \includegraphics[width=0.24\textwidth]{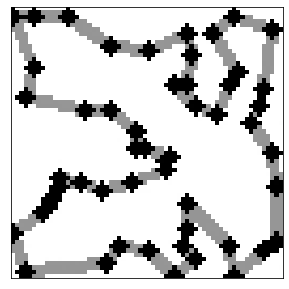}
    \includegraphics[width=0.24\textwidth]{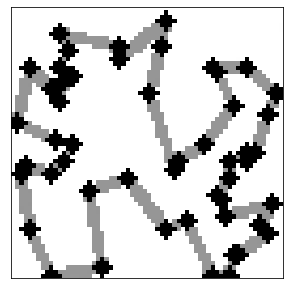}
    \caption{Two unconditional samples from the diffusion model trained on images of solved TSPs.}
    \label{fig:tsp_gen}
\end{wrapfigure}

\paragraph{Diffusion model training.}
We use a dataset of Euclidean TSPs, with ground truth tours obtained by a state-of-the-art TSP solver \cite{concorde}, from \cite{kool2019attention} (we consider two variants of the dataset, each with $\sim$1.5m training graphs: with 50 vertices in each graph and with a varying number from 20 to 50 vertices in each graph). Each training tour is represented via its adjacency matrix $A$ and encoded as an image $f(A)$. We then train a DDPM with the U-Net architecture from \cite{dhariwal2021diffusion} on all of such encoded image. Model and training details can be found in the Appendix (\S\ref{sec:tsp_appendix}). Some unconditional samples from the trained DDPM are shown in Fig.~\ref{fig:tsp_gen}; most samples indeed resemble image representations of tours.

\begin{figure}[t]
    \centering
    \begin{tabular}{@{}c@{}c@{}c@{}c@{}c@{}c@{}c@{\hskip1pt}c@{\hskip1pt}c@{}}
    &\multicolumn{5}{c}{\small Optimize latent adjacency matrix w.r.t.\ denoising model}
    &\multicolumn{2}{c}{\small Recover tour}
    \\\cmidrule(lr){2-6}\cmidrule(lr){7-8}
        \small Input& \small $t=$256 & \small $t=$192& \small $t=$128 & \small $t=$64&\small $t=$0 & \small Extracted & + 2-opt &  \small Oracle\\
    \includegraphics[width=0.11\textwidth,trim=0 0 0 0,clip]{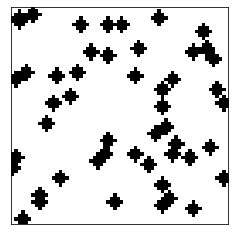}&
    \includegraphics[width=0.11\textwidth,trim=0 0 0 0,clip]{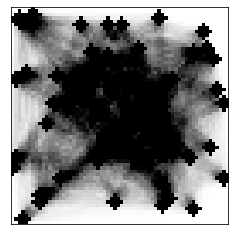}&
    \includegraphics[width=0.11\textwidth,trim=0 0 0 0,clip]{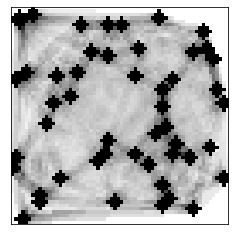}&
    \includegraphics[width=0.11\textwidth,trim=0 0 0 0,clip]{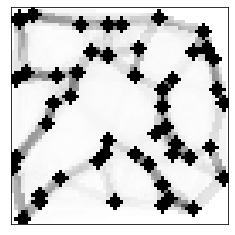}&
    \includegraphics[width=0.11\textwidth,trim=0 0 0 0,clip]{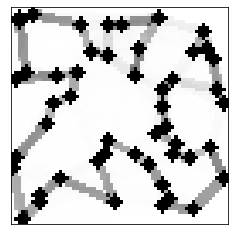}&
    \includegraphics[width=0.11\textwidth,trim=0 0 0 0,clip]{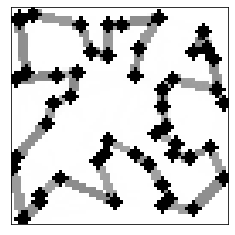}&
    \includegraphics[width=0.11\textwidth,trim=5 0 5 0,clip]{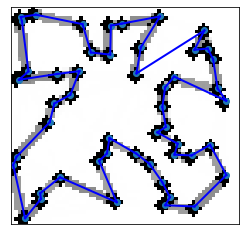}&
    \includegraphics[width=0.11\textwidth,trim=5 0 5 0,clip]{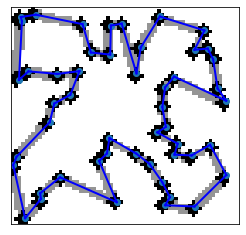}&
    \includegraphics[width=0.11\textwidth,trim=5 0 5 0,clip]{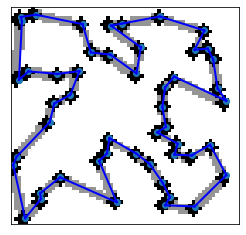}\\[-2pt]
    \includegraphics[width=0.11\textwidth,trim=0 0 0 0,clip]{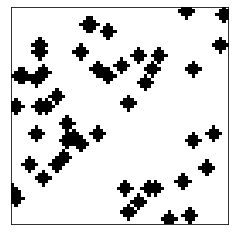}&
    \includegraphics[width=0.11\textwidth,trim=0 0 0 0,clip]{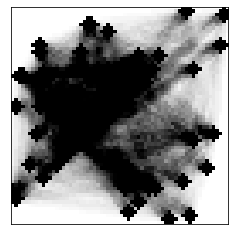}&
    \includegraphics[width=0.11\textwidth,trim=0 0 0 0,clip]{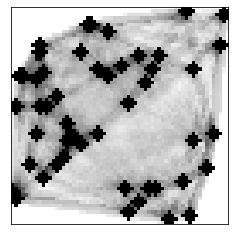}&
    \includegraphics[width=0.11\textwidth,trim=0 0 0 0,clip]{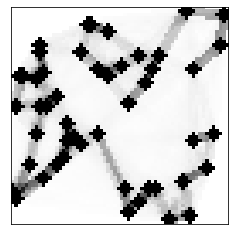}&
    \includegraphics[width=0.11\textwidth,trim=0 0 0 0,clip]{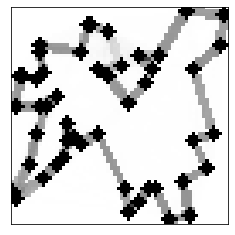}&
    \includegraphics[width=0.11\textwidth,trim=0 0 0 0,clip]{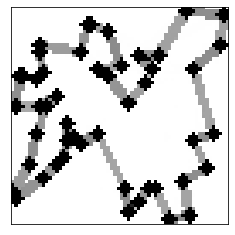}&
    \includegraphics[width=0.11\textwidth,height=0.11\textwidth,trim=5 0 5 0,clip]{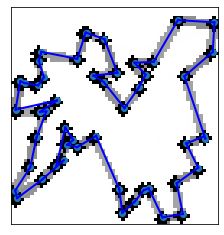}&
    \includegraphics[width=0.11\textwidth,height=0.11\textwidth,trim=5 0 5 0,clip]{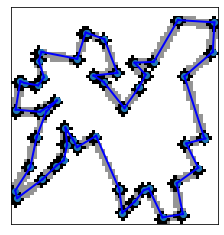}&
    \includegraphics[width=0.11\textwidth,height=0.11\textwidth,trim=5 0 5 0,clip]{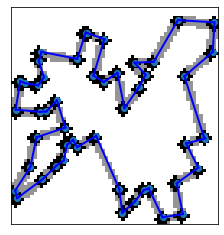}\\
    \end{tabular}
    \caption{The procedure for solving the Euclidean TSP with a DDPM: Gradient descent is performed on a latent adjacency matrix $A$ to minimize a stochastic denoising loss on an \emph{image representation} $f(A)$ with steadily decreasing amounts of noise (here, 256 steps). In the process, pieces of the tour are `burned in' and later recombined in creative ways. Finally, a tour is extracted from the inferred adjacency matrix and refined by uncrossing moves. For both problems shown, the length of the inferred tour is within 1\% of the optimum.}
    \label{fig:tsp_latent}
\end{figure}

\paragraph{Solving new TSPs.}
Suppose we are given a new set of points $v_1,\dots,v_N$. Solving the TSP requires finding the adjacency matrix $A$ of a tour of minimal length. As a differentiable relaxation, we set $A=S+S^\top$, where $S$ is a stochastic $N\times N$ matrix with zero diagonal (parametrized via softmax of a matrix of parameters over rows). We run the inference procedure using the trained DDPM $p_{\rm DDPM}(f(A))$ as a prior to estimate $A$. The hyperparameters and noise schedule are described in \S\ref{sec:tsp_appendix}. Examples of the optimization are shown in Fig.~\ref{fig:tsp_latent}.

Although the inferred $A$ is usually sharp (i.e., all entries close to 0 or 1), rounding $A$ to 0 or 1 does not always give the adjacency matrix of a tour (see, for example, the top row of Fig.~\ref{fig:tsp_latent}; other common incorrect outputs include pairs of disjoint tours). To extract a tour from the inferred $A$, we greedily insert edges to form an initial proposal, then refine it using a standard and lightweight combinatorial procedure, the 2-opt heuristic \cite{lkh} (amounting to iteratively uncrossing pairs of edges that intersect). The entire procedure is shown in Fig.~\ref{fig:tsp_latent}, and full details can be found in the Appendix (\S\ref{sec:tsp_appendix}).

\begin{table}[t]
    \caption{\textit{Left:} Mean tour length and optimality gap on Euclidean TSP test sets. The baseline results from \cite{kool2019attention,joshi2019efficient,bresson2021tsptransformer} are taken from the respective papers. The two DDPMs were trained on 1.5m images of solved TSP instances (with different numbers of vertices) and used to infer latent adjacency matrices in the test set. \textit{Right:} Performance of the DDPM trained on images of 50-vertex TSP instances with different numbers of inference steps (see the Appendix (\S\ref{sec:tsp_appendix}) for time schedule details). We also show the mean number of 2-opt (uncrossing) steps per instance, suggesting that the DDPM prior assigns high likelihood to adjacency matrices that are in less need of refinement.}
    \label{tab:tsp}
    \centering
    \resizebox{0.49\textwidth}{!}{
    \begin{tabular}{lcccc}
\toprule
&\multicolumn{2}{c}{$N=50$}&\multicolumn{2}{c}{$N=100$}\\
\cmidrule(lr){2-3}\cmidrule(lr){4-5}
Algorithm&Obj&Gap \%&Obj&Gap \%\\\midrule
Oracle (Concorde \cite{concorde})  & 5.69 & 0.00 & 7.759 & 0.00 \\\midrule
2-opt \cite{lkh} & 5.86 & 2.95 & 8.03 & 3.54 \\\midrule
Transformer \cite{kool2019attention} & 5.80 & 1.76 & 8.12 & 4.53 \\
GNN \cite{joshi2019efficient} &  5.87 & 3.10 & 8.41  & 8.38 \\
Transformer \cite{bresson2021tsptransformer} & 5.71 & 0.31 & 7.88 & 1.42 \\
\midrule
Diffusion 20--50 & 5.76 & 1.23 & 7.92 & 2.11 \\
Diffusion 50 & 5.76 & 1.28 & 7.93 & 2.19 \\
\bottomrule
\end{tabular}
    }
    \hfill
    \resizebox{0.49\textwidth}{!}{
    \begin{tabular}{lcccccc}
\toprule
&\multicolumn{3}{c}{$N=50$}&\multicolumn{3}{c}{$N=100$}\\
\cmidrule(lr){2-4}\cmidrule(lr){5-7}
Diff.\ steps&Obj&Gap \%&Steps&Obj&Gap \%&Steps\\\midrule
256 & 5.763 & 1.28 & 11.6 & 7.930 & 2.19 & 50.6 \\
64 & 5.780 & 2.60 & 14.3 & 7.942 & 2.35 & 45.7 \\
16 & 5.858 & 2.98 & 25.9 & 8.052 & 3.78 & 58.6 \\
4 & 5.851 & 2.86 & 23.9 & 8.031 & 3.50 & 52.8 \\ \midrule
2-opt & 5.856 & 2.95 & 24.4 & 8.034 & 3.54 & 53.0 \\
\bottomrule
\end{tabular}
    }
\end{table}

\paragraph{Results.} We evaluate the trained models on test sets of 1280 graphs each with $N=50$ and $N=100$ vertices. We report the average length of the inferred tour and the gap (discrepancy from the length of the ground truth tour) in Table~\ref{tab:tsp} (left), from which we make several observations.
\begin{itemize}[leftmargin=*,itemsep=0pt,topsep=0pt]
    \item The right side of Table~\ref{tab:tsp} shows the number of 2-opt (edge uncrossing) steps performed in the refinement step of the algorithm when the inference algorithm is run for varying numbers of steps. Running the inference with more steps results in extracted tours that are closer to local minima with respect to the 2-opt neighbourhood, indicating that the DDPM encodes meaningful information about the shape of tours.
    \item The DDPM inference is competitive with recent baseline algorithms that do not use beam search in generation of the tour (those shown in the table). These baseline algorithms improve when beam search decoding with very large beam size is used, but encounter diminishing returns as the computation cost grows. Our performance on the 100-vertex problems is similar to \cite{kool2019attention} with the largest beam size they report (5000), which has 2.18\% gap, while having similar computation time.
    \item The model trained on problems with 50 nodes performs almost identically to the model trained on problems with 50 or fewer nodes, and both models generalize better than baseline methods from 50-node problems to the out-of-distribution 100-node problems. 
\end{itemize}
We emphasize a unique feature of our algorithm: all `reasoning' in our inference procedure happens via the image space. This property also leads to sublinear computation cost scaling with increasing size of the graph -- as long as it can reasonably be represented in a $64\times64$ image -- since most of the computation cost of inference is borne by running the denoiser on images of a fixed size. In the Appendix (\S\ref{sec:tsp_appendix}) we explore the generalization of the model trained on 20- to 50-node TSP instances to problems with 200 nodes and discuss potential extensions.

\section{Conclusion}
\label{sec:conclusion}
We have shown how inference in denoising diffusion models can be performed under constraints in a variety of settings. Imposing constraints that arise from pretrained classifiers enables conditional generation, while common-sense conditions, such as mutual information with a clustering or divergence from weak labels, can lead to models that are less sensitive to domain shift in the distribution of conditioning data.

A notable limitation of DDPMs, which is inherited by our algorithms, is the high cost of inference, requiring a large number of passes through the denoising network to generate a sample. We expect that with further research on DDPMs for which inference procedures converge in fewer steps \cite{salimans2022progressive,xiao2022tackling}, plug-and-play use of DDPMs will become more appealing in various applications.

Finally, our results on the traveling salesman problem illustrate the ability of DDPMs to reason over uncertain hypotheses in a manner that can mimic human `puzzle-solving' behavior. These results open the door to future research on using DDPMs to efficiently generate candidates in combinatorial search problems.

\section*{Acknowledgments}

The authors thank the anonymous NeurIPS 2022 reviewers for their comments. 

All authors are funded by their primary institutions. Partial support was provided by the NASA Biodiversity program (Award 80NSSC21K1027),  NSF grants IIS-2123920 and  IIS-2212046, and the Partner University Fund 4D Vision award.

\bibliographystyle{plainnat}
\bibliography{references}

\section*{Checklist}

\begin{enumerate}

\item For all authors...
\begin{enumerate}
  \item Do the main claims made in the abstract and introduction accurately reflect the paper's contributions and scope?
    \answerYes{}
  \item Did you describe the limitations of your work?
    \answerYes{See the conclusion and discussion throughout the paper.}
  \item Did you discuss any potential negative societal impacts of your work?
    \answerNA{Although no immediate negative societal impacts are expected, researchers should bear in mind the risks of flexible conditional generation of images, e.g., for creating `deep fakes'.}
  \item Have you read the ethics review guidelines and ensured that your paper conforms to them?
    \answerYes{}
\end{enumerate}

\item If you are including theoretical results...
\begin{enumerate}
  \item Did you state the full set of assumptions of all theoretical results?
    \answerNA{}
        \item Did you include complete proofs of all theoretical results?
    \answerNA{}
\end{enumerate}

\item If you ran experiments...
\begin{enumerate}
  \item Did you include the code, data, and instructions needed to reproduce the main experimental results (either in the supplemental material or as a URL)?
    \answerYes{For most experiments; see the Appendix.}
  \item Did you specify all the training details (e.g., data splits, hyperparameters, how they were chosen)?
    \answerYes{See the Appendix and relevant experiment sections.}
        \item Did you report error bars (e.g., with respect to the random seed after running experiments multiple times)?
    \answerNo{Main experiments were run one time.}
        \item Did you include the total amount of compute and the type of resources used (e.g., type of GPUs, internal cluster, or cloud provider)?
    \answerYes{See the Appendix.}
\end{enumerate}

\item If you are using existing assets (e.g., code, data, models) or curating/releasing new assets...
\begin{enumerate}
  \item If your work uses existing assets, did you cite the creators?
    \answerYes{See the relevant experiment sections.}
  \item Did you mention the license of the assets?
    \answerNo{But all datasets used are free to use for research purposes; see the relevant citations.}
  \item Did you include any new assets either in the supplemental material or as a URL?
    \answerNA{}
  \item Did you discuss whether and how consent was obtained from people whose data you're using/curating?
    \answerNA{}
  \item Did you discuss whether the data you are using/curating contains personally identifiable information or offensive content?
    \answerNA{}
\end{enumerate}

\item If you used crowdsourcing or conducted research with human subjects...
\begin{enumerate}
  \item Did you include the full text of instructions given to participants and screenshots, if applicable?
    \answerNA{}
  \item Did you describe any potential participant risks, with links to Institutional Review Board (IRB) approvals, if applicable?
    \answerNA{}
  \item Did you include the estimated hourly wage paid to participants and the total amount spent on participant compensation?
    \answerNA{}
\end{enumerate}

\end{enumerate}

\newpage
\appendix

\counterwithin{figure}{section}
\counterwithin{table}{section}

\section{Deriving a Gaussian approximation to the posterior}
\label{sec:gaussian_estimate}

We repeat the derivation in \S\ref{sec:ddpm_prior} for a Gaussian, rather than a point estimate of the posterior.

Recall that if $p(\vect{x})$ is the likelihood (\ref{eq:marginal}) of $\vect{x}$ under a DDPM, then in the first expectation of (\ref{eq:F}) we should use $q(\vect{h}=\{\vect{x}_T,...,\vect{x}_1\}|\vect{x}_0=\vect{x})=\prod_{t=1}^T  q(\vect{x}_t \mid \vect{x}_{t-1})$. A computationally and notationally convenient form for the approximate posterior over $\vect{x}=\vect{x}_0$ is a scalar-covariance Gaussian:
\begin{equation}
    q(\vect{x})= \normal(\vect{x} ; \sqrt{1-\psi}\eta, \psi \mat{I}).
    \label{eq:q_def_gaussian}
\end{equation}
We can sample $\vect{x}_t$ at any arbitrary time step  as
\begin{equation}
    q(\vect{x}_t) = \normal(\vect{x}_t ; \sqrt{\bar{\alpha}_t}\vect{\eta}, (1-\bar{\alpha}_t) \mat{I})
    \label{eq:diff_sample_gaussian}
\end{equation}
where $\alpha_t = 1 - \beta_t$ and $\bar{\alpha}_t = \prod_{i=0}^t \alpha_t$, with the convention that $\beta_{0}=\psi$ (note the difference with (\ref{eq:diff_sample}), which is the special case of (\ref{eq:diff_sample_gaussian}) with $\beta_0=0$). Analogously to \cite{ho2020ddpm}, we can also extract a conditional Gaussian  $q(\vect{x}_{t-1} \mid \vect{x}_t, \psi, \eta)$ and express the first expectation in (\ref{eq:F}) as
\begin{equation}
    -\expect_{q(\vect{x})q(\vect{h}\mid \vect{x})} [\log p(\vect{x},\vect{h})  - \log q(\vect{x})q(\vect{h}|\vect{x})] = \sum_t \KL(q(\vect{x}_{t-1} \mid \vect{x}_t, \eta, \psi) \,\|\, p_{\theta}(\vect{x}_{t-1} \mid \vect{x}_t)),
    \label{eq:sum_of_KL_gaussian}
\end{equation}
which after reparametrization \cite{ho2020ddpm} leads to
\begin{equation}
    \sum_t w_t(\beta)\expect_{\boldsymbol{\epsilon} \sim \normal (\vect{0},\mat{I})}
    [\norm{\boldsymbol{\epsilon} - \boldsymbol{\epsilon}_{\theta}(\vect{x}_t,t)}_2^2],\quad 
    \vect{x}_t = \sqrt{\bar{\alpha}_t}\vect{\eta} + \sqrt{1-\bar{\alpha}_t}\boldsymbol{\epsilon}
    \label{eq:reparametrization_gaussian}
\end{equation}
where the link between the stage $t$ noise reconstruction $\boldsymbol{\epsilon}_{\theta}(\vect{x}_t,t)$ and the model's expectation $\boldsymbol{\mu}_{\theta}(\vect{x}_t,t)$ is 
\begin{equation}
    \boldsymbol{\mu}_\theta(\vect{x}_t,t)=\frac{1}{\sqrt{\alpha_t}}\left(\vect{x}_t-\frac{\beta_t}{\sqrt{1-\bar\alpha_t}}\boldsymbol{\epsilon}_\theta(\vect{x}_t,t)\right).
    \label{eq:reparametrization_mu_gaussian}
\end{equation}
Assuming uniform weighting of the noising steps as before, the free energy in (\ref{eq:F}) reduces to
\begin{equation}
    F= \sum_t \expect_{\epsilon \sim \normal (\vect{0},\mat{I})}
    [\norm{\boldsymbol{\epsilon} - \boldsymbol{\epsilon}_{\theta}(\vect{x}_t,t)}_2^2]-\expect_{q(\vect{x})} [\log c(\vect{x},\vect{y})].
    \label{eq:F_final_gaussian}
\end{equation}
Unlike (\ref{eq:dirac_opt}), (\ref{eq:F_final_gaussian}) involves an expectation over a Gaussian variable. To optimize through this expectation, one could use the reparametrization trick: $\expect_{q(\vect{x})} [\log c(\vect{x},\vect{y})]=\expect_{\boldsymbol{\epsilon}_q \sim \normal(0,\mat{I})} [\log c(\sqrt{1-\psi}\boldsymbol{\eta}+\boldsymbol{\epsilon}_q \sqrt{\psi},\vect{y})]$.

\section{Experiment details and extensions}

\subsection{MNIST}
\label{sec:mnist_appendix}
\paragraph{Training the DDPM.} To train the diffusion model we used the U-Net architecture of \cite{dhariwal2021diffusion} with a linear $\beta_t$ schedule and $T=1000$ diffusion steps. We trained the network for 10 epochs, with a batch size of 128 samples, using the Adam optimizer and a learning rate of $10^{-4}$.

\begin{figure}[t]
\centering
    \begin{tabular}{cc}
        \includegraphics[width=0.45\textwidth]{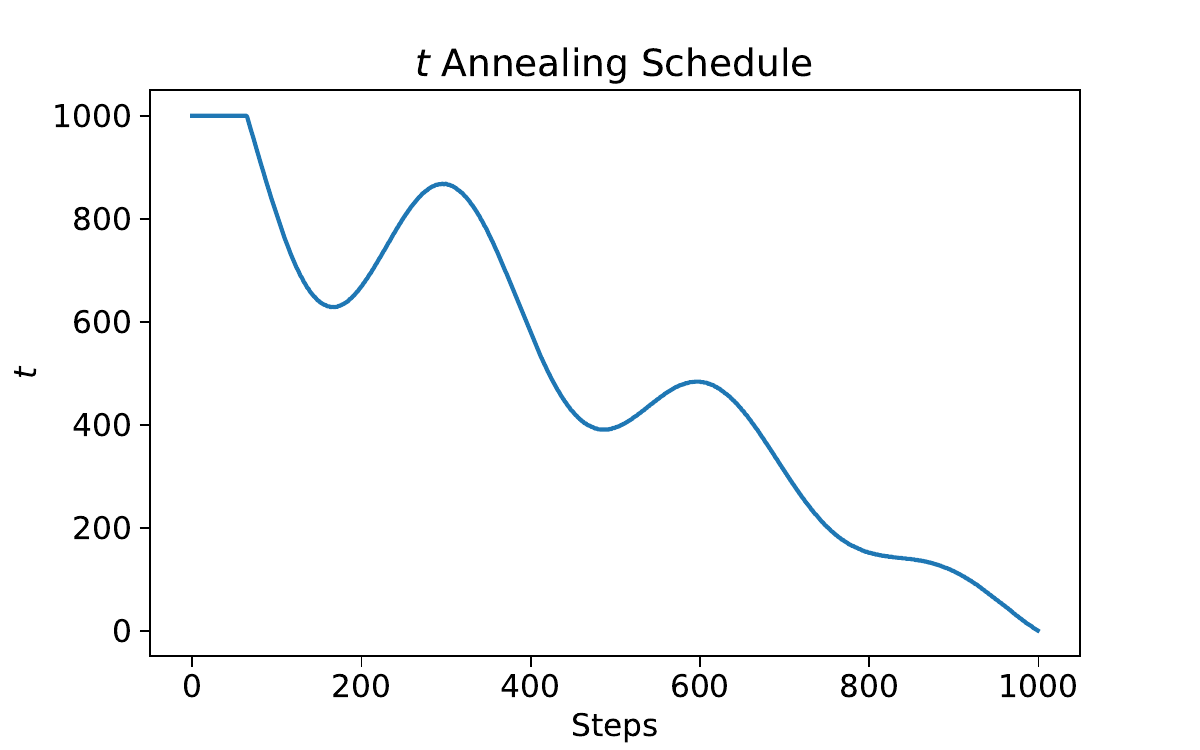} &
        \includegraphics[width=0.45\textwidth]{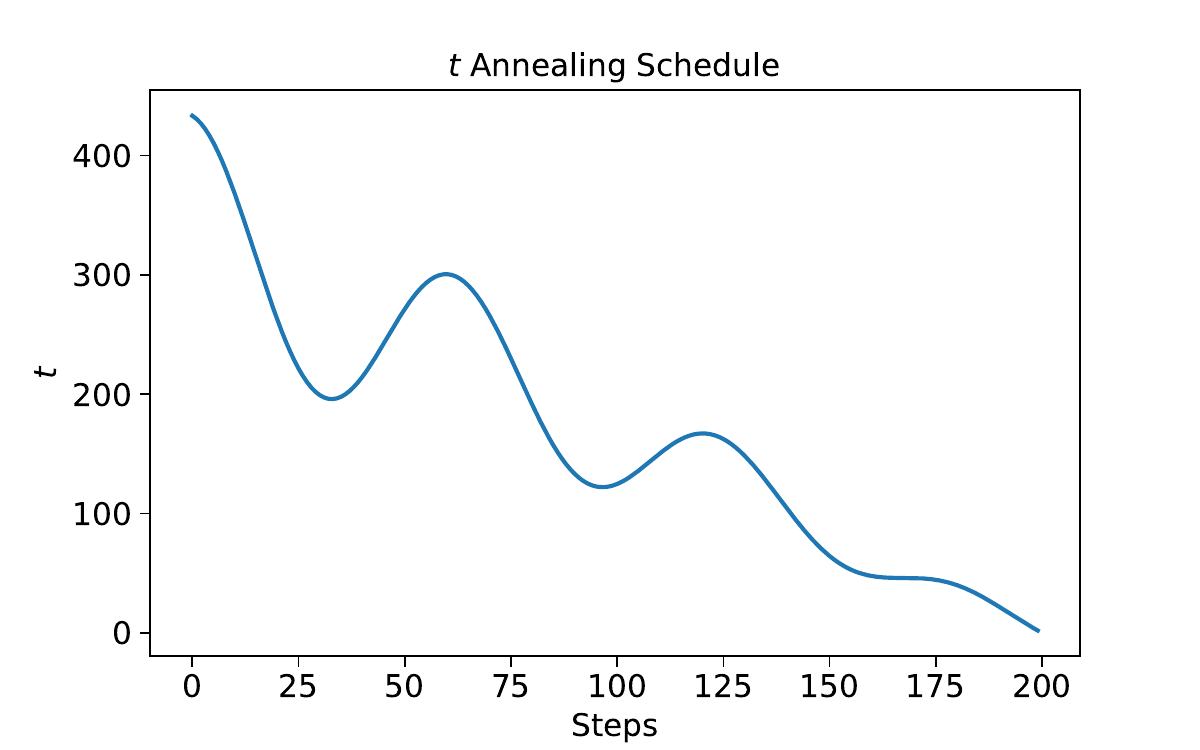} \\
        (a) & (b) \\
    \end{tabular}
    \caption{Inference $t$ annealing schedules for the (a) MNIST (b) Land Cover experiments. We do not necessarily need to optimize for all $T=1000$ values to generate samples, as shown in (b). The TSP and FFHQ experiments use similarly defined schedules.}
    \label{fig:annealing_schedules}
\end{figure}

\paragraph{Performing inference.} For all inference examples, we performed 1000 optimization steps with the Adam optimizer and a learning rate $10^{-2}$. We employed a cosine-modulated, linearly decreasing $t$ annealing schedule as shown in Fig.~\ref{fig:annealing_schedules} (a). We empirically designed this annealing process following the observation that the linearly decreasing $t$ values guide the inference procedure in a coarse-to-fine manner that starts by deciding the overall structure of the inferred sample and then adding in details. We also added the oscillating component to allow for revisions of the coarser structures that are to be made after having inferred specific details.

When performing the optimization step of Algorithm 1, we observed that it was important to gradually reduce the effect of the condition $c(\vect{x},\vect{y})$ in order to obtain good sample quality. In practice, we linearly decreased the weight of the conditional component of the loss, from $\lambda_T = 10^{-2}$ to $\lambda_1 = 0$ as we performed the optimization steps from $T,\dots,1$. This can be attributed to the fact that the conditions we used provide guidance for the steps made at larger values of $t$, where the shape and orientation of a digit are decided. When combining two or more conditions the weighting is applied to all of them.

\subsection{FFHQ}
\label{sec:ffhq_appendix}
\paragraph{Performing inference.} For the conditional generation experiments on the FFHQ dataset we utilized the pretrained DDPM model provided by \cite{baranchuk2022labelefficient}. The face attribute classifier network was a ResNet-18 network trained on the face attributes given in the CelebA dataset \cite{liu2015faceattributes}. To run our inference algorithm we performed 200 optimization steps with the Adamax optimizer, choosing $(\beta_1,\beta_2)=(0.9,0.999)$ and a linearly decreasing learning rate from $1$ to $0.5$. The $t$ annealing schedule was similar to the one used for the land cover segmentation experiments (Fig.~\ref{fig:annealing_schedules} (b)) but for $t$ values now ranging from 1000 to 200. Additionally, in this experiment we found that balancing between the diffusion and auxiliary losses with a carefully chosen weighting term was difficult. Thus, we opted for a different approach where we clipped the gradient norm of the auxiliary objective to $\frac{1}{2}$ of the gradient norm of the diffusion denoising loss. 

\begin{figure}[t]
    \centering
    \small
    \begin{tabular}{@{}c@{\hspace{1mm}}@{\hspace{1mm}}c@{\hspace{2mm}}c@{\hspace{2mm}}c@{\hspace{2mm}}c@{\hspace{2mm}}c@{}}
        \includegraphics[width=0.15\textwidth]{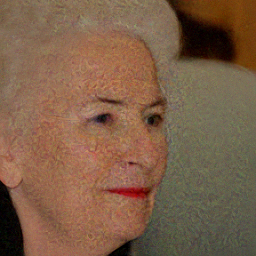} &
        \includegraphics[width=0.15\textwidth]{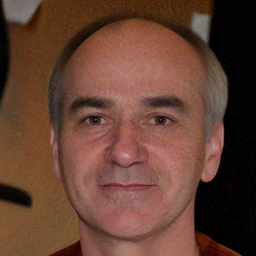} &
        \includegraphics[width=0.15\textwidth]{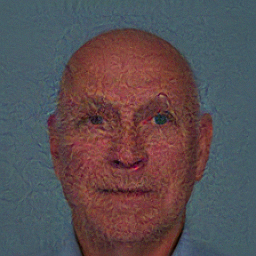} &
        \includegraphics[width=0.15\textwidth]{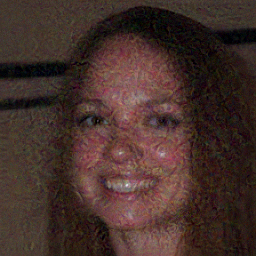} &
        \includegraphics[width=0.15\textwidth]{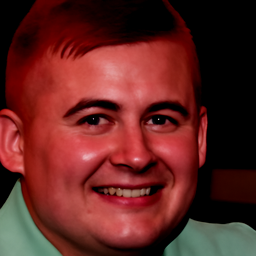} &
        \includegraphics[width=0.15\textwidth]{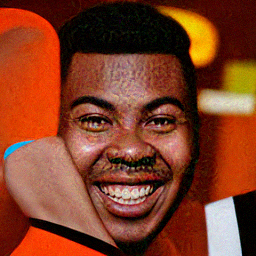} \\
        \includegraphics[width=0.15\textwidth]{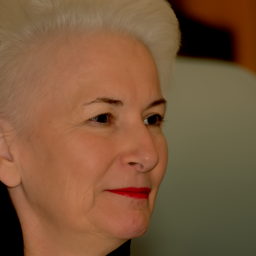} &
        \includegraphics[width=0.15\textwidth]{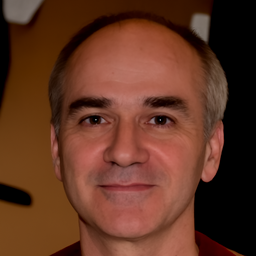} &
        \includegraphics[width=0.15\textwidth]{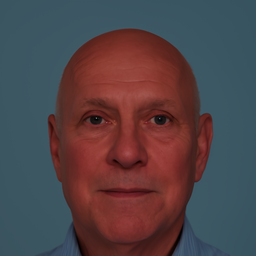} &
        \includegraphics[width=0.15\textwidth]{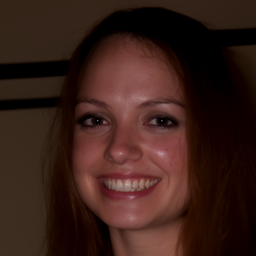} &
        \includegraphics[width=0.15\textwidth]{figs/faces/male_rosy_cheeks_0_finetuned_200.png} &
        \includegraphics[width=0.15\textwidth]{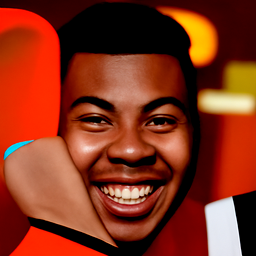} \\
        Not Male & Male & Not Young & Young & Male \& & Male \& Smiling \\
        & & & & Rosy Cheeks & \& Mustache \\
        \multicolumn{6}{c}{(a)} \\
        \\
        \includegraphics[width=0.15\textwidth]{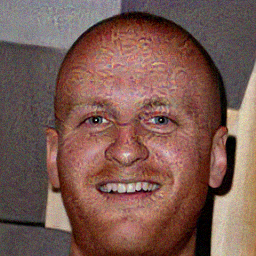} &
        \includegraphics[width=0.15\textwidth]{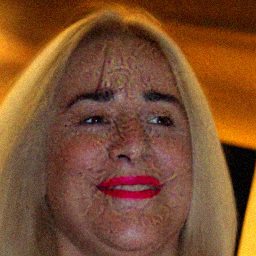} &
        \includegraphics[width=0.15\textwidth]{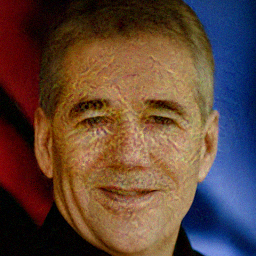} &
        \includegraphics[width=0.15\textwidth]{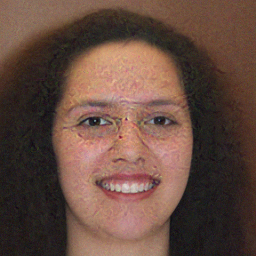} &
        \includegraphics[width=0.15\textwidth]{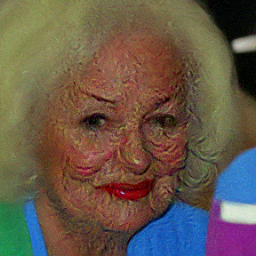} &
        \includegraphics[width=0.15\textwidth]{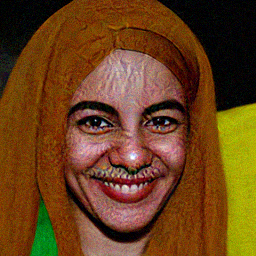} \\
        \includegraphics[width=0.15\textwidth]{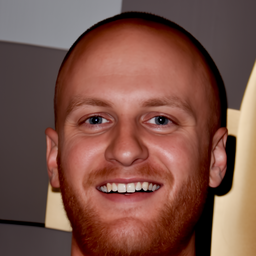} &
        \includegraphics[width=0.15\textwidth]{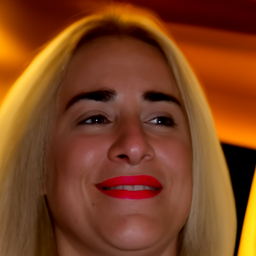} &
        \includegraphics[width=0.15\textwidth]{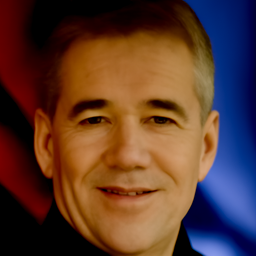} &
        \includegraphics[width=0.15\textwidth]{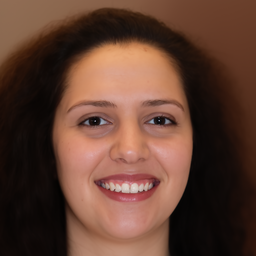} &
        \includegraphics[width=0.15\textwidth]{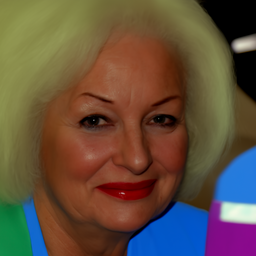} &
        \includegraphics[width=0.15\textwidth]{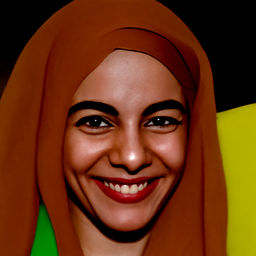} \\
        Not Male \& & Not Male \& & Blonde \& & Eyeglasses & Not Male \& & Smiling \& \\
        Bald & Beard & Black Hair & & Wavy Hair & Mustache \\
        \multicolumn{6}{c}{(b)}
    \end{tabular}
    \caption{(a) Additional conditional samples $\vect{x}$ for constraints $c(\vect{x},\vect{y})$ with various attribute sets $\vect{y}$. (b) Failure cases of conditional generation with their attribute sets $\vect{y}$. For both sets of images we show the inference results (top) and the image after denoising as in \cite{nie2022diffusion} to remove artifacts that appear due to optimizing the classifier constraint (bottom).}
    \label{fig:face_samples_appendix}
\end{figure}

\paragraph{Further discussion on samples.} In Fig.~\ref{fig:face_samples_appendix} we demonstrate additional conditionally-generated samples from the unconditional DDPM and the attribute classifier. In the first set of examples we show that although we may find modes of $p_{DDPM}(x)$ that satisfy to a level the condition $\vect{y}$ set by the classifier, the sample quality is not always on-par with unconditionally generated samples, like those presented in \cite{baranchuk2022labelefficient}. We can attribute that to the fact that for natural images, in contrast to segmentation labels, the mode may not always be a good-looking sample from the distribution. Our method to mitigate that, along with the classifier noise artifacts left from the optimization process, is to run the diffusion denoising procedure starting from a low temperature $t$. Although this may improve the visual quality of the result, in some cases our choice of $t$ is not large enough to move the sample far enough from the inferred $\vect{x}$. If we choose a larger $t$ however we risk erasing the attributes we aimed to generate in the first place.

In the second set of samples, we first show how conflicting attributes are resolved. When the constraint is set to satisfy two attributes that contradict each other we observed that the inferred sample $\vect{x}$ tends to gravitate towards a single randomly-chosen direction. This is evident in the first two examples where we set the \textit{not male} attribute along with a male-correlated attribute. In each of them only a single condition, either the not male or the male-related, is satisfied. In the \textit{blonde}+\textit{black hair} example we could argue that a mix of the two attributes is present in the inferred sample. However, the classifier predictions for that specific image tell us that the person shown is exclusively blonde.

We also show a set of failure cases where the classifier `painted' the features related to the desired attribute but the diffusion prior did not complete the sample in a correct way. For instance, in the \textit{eyeglasses} example we see that the classifier has drawn an outline of the eyeglass edges on the generated face but the diffusion model has failed to pick up the cue. Similarly, when asking for \textit{wavy hair} we see curves that can fool the classifier into thinking that the person has curly hair, or when the attributes set are \textit{smiling}+\textit{mustache} we observe a comically drawn mustache on the generated face. Since the conditioning depends both on the diffusion prior and the robustness of the classifier we believe that with better classifier training we could improve the result in such cases.

\subsection{Land cover}
\label{sec:land_cover_appendix}
\begin{figure}[t]
    \centering
    \begin{tabular}{ccccc}
        \includegraphics[width=0.15\textwidth]{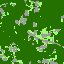} &
        \includegraphics[width=0.15\textwidth]{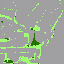} &
        \includegraphics[width=0.15\textwidth]{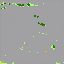} &
        \includegraphics[width=0.15\textwidth]{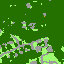} &
        \includegraphics[width=0.15\textwidth]{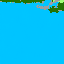} \\
        \includegraphics[width=0.15\textwidth]{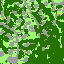} &
        \includegraphics[width=0.15\textwidth]{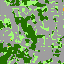} &
        \includegraphics[width=0.15\textwidth]{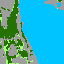} &
        \includegraphics[width=0.15\textwidth]{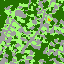} &
        \includegraphics[width=0.15\textwidth]{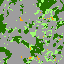} \\
        \includegraphics[width=0.15\textwidth]{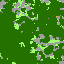} &
        \includegraphics[width=0.15\textwidth]{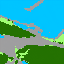} &
        \includegraphics[width=0.15\textwidth]{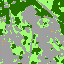} &
        \includegraphics[width=0.15\textwidth]{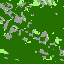} &
        \includegraphics[width=0.15\textwidth]{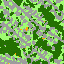} \\
        \includegraphics[width=0.15\textwidth]{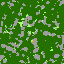} &
        \includegraphics[width=0.15\textwidth]{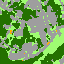} &
        \includegraphics[width=0.15\textwidth]{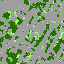} &
        \includegraphics[width=0.15\textwidth]{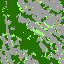} &
        \includegraphics[width=0.15\textwidth]{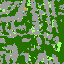} \\
        \includegraphics[width=0.15\textwidth]{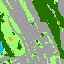} &
        \includegraphics[width=0.15\textwidth]{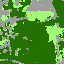} &
        \includegraphics[width=0.15\textwidth]{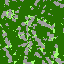} &
        \includegraphics[width=0.15\textwidth]{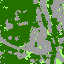} &
        \includegraphics[width=0.15\textwidth]{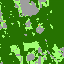} \\
        \includegraphics[width=0.15\textwidth]{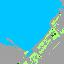} &
        \includegraphics[width=0.15\textwidth]{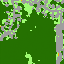} &
        \includegraphics[width=0.15\textwidth]{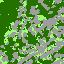} &
        \includegraphics[width=0.15\textwidth]{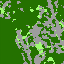} &
        \includegraphics[width=0.15\textwidth]{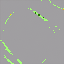} \\
    \end{tabular} \\
    {\footnotesize
    \legendbox{water} Water
    \legendbox{impervious} Impervious Surface
    \legendbox{barren} Soil and Barren
    \legendbox{trees} Trees and Forest
    \legendbox{grass} Grass and Herbaceous}
    \caption{Unconditional samples from the DDPM trained on land cover segmentations.}
    \label{fig:seg_samples_appendix}
\end{figure}

\paragraph{Training the DDPM.} The land cover DDPM was trained on $\frac{1}{4}$-resolution, $64\times64$ patches of land cover labels, randomly sampled from the Pittsburgh, PA tiles of the EnviroAtlas dataset \cite{pickard2015enviroatlas}. For the diffusion network, we used the U-Net architecture of \cite{dhariwal2021diffusion}, a linear $\beta_t$ schedule and $T=1000$ diffusion steps. We trained with $10^5$ batches of size 32, using the Adam optimizer and a learning rate of $10^{-4}$. Additional samples from the unconditional diffusion model are shown in Fig.~\ref{fig:seg_samples_appendix}. We observe that the model has learned both structures that are independent of the geography, such as the continuity of roads and the suburban building planning, and PA-specific ones, such as buildings nested in forested areas, which may not be as common in AZ for instance.

\begin{figure}
    \centering
    \small
    \begin{tabular}{ccccc}
        & & Weak & & \\
        Image & Clustering $\vect{z}$ & Labels $\ell_{\rm weak}$ & Inferred $\vect{x}$ & Ground Truth \\
        \includegraphics[width=0.15\textwidth]{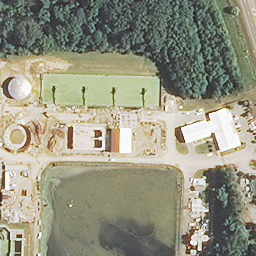} &
        \includegraphics[width=0.15\textwidth]{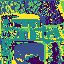} &
        \includegraphics[width=0.15\textwidth]{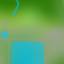} & \includegraphics[width=0.15\textwidth]{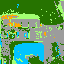} &
        \includegraphics[width=0.15\textwidth]{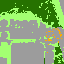} \\
        \includegraphics[width=0.15\textwidth]{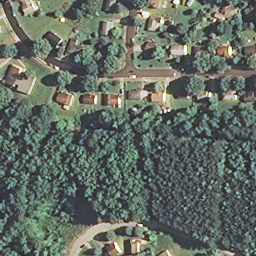} &
        \includegraphics[width=0.15\textwidth]{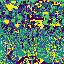} &
        \includegraphics[width=0.15\textwidth]{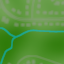} & \includegraphics[width=0.15\textwidth]{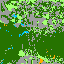} &
        \includegraphics[width=0.15\textwidth]{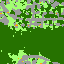} \\
        \includegraphics[width=0.15\textwidth]{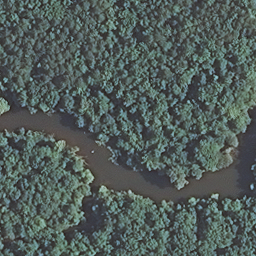} &
        \includegraphics[width=0.15\textwidth]{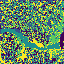} &
        \includegraphics[width=0.15\textwidth]{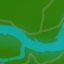} & \includegraphics[width=0.15\textwidth]{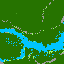} &
        \includegraphics[width=0.15\textwidth]{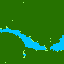} \\
        \includegraphics[width=0.15\textwidth]{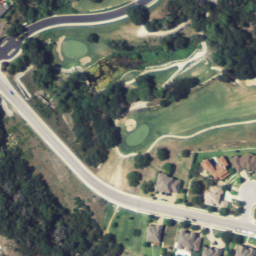} &
        \includegraphics[width=0.15\textwidth]{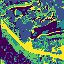} &
        \includegraphics[width=0.15\textwidth]{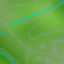} & \includegraphics[width=0.15\textwidth]{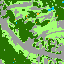} &
        \includegraphics[width=0.15\textwidth]{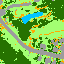} \\
        \includegraphics[width=0.15\textwidth]{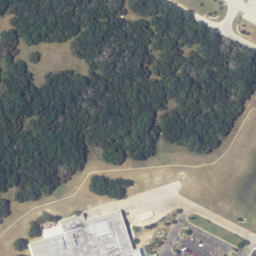} &
        \includegraphics[width=0.15\textwidth]{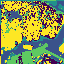} &
        \includegraphics[width=0.15\textwidth]{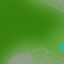} & \includegraphics[width=0.15\textwidth]{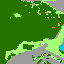} &
        \includegraphics[width=0.15\textwidth]{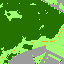} \\
        \includegraphics[width=0.15\textwidth]{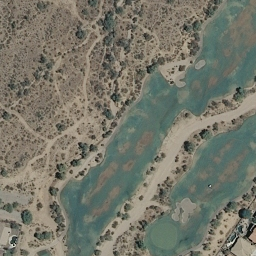} &
        \includegraphics[width=0.15\textwidth]{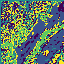} &
        \includegraphics[width=0.15\textwidth]{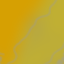} & \includegraphics[width=0.15\textwidth]{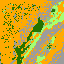} &
        \includegraphics[width=0.15\textwidth]{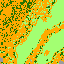} \\
        \includegraphics[width=0.15\textwidth]{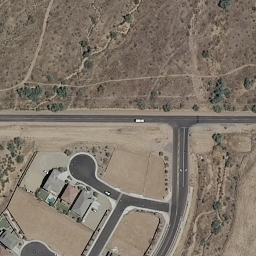} &
        \includegraphics[width=0.15\textwidth]{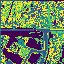} &
        \includegraphics[width=0.15\textwidth]{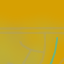} & \includegraphics[width=0.15\textwidth]{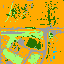} &
        \includegraphics[width=0.15\textwidth]{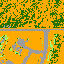} \\
    \end{tabular} \\
    {\footnotesize
    \legendbox{water} Water
    \legendbox{impervious} Impervious Surface
    \legendbox{barren} Soil and Barren
    \legendbox{trees} Trees and Forest
    \legendbox{grass} Grass and Herbaceous}
    \caption{Segmentation inference results.}
    \label{fig:seg_inference_samples_appendix}
\end{figure}

\paragraph{Performing inference.} Since we initialize the inference procedure with the weak labels we require fewer optimization steps and do not have to start the search from $t=1000$. Thus, to infer the land cover segmentations we only perform 200 optimization steps using the Adam optimizer, with a linearly decreasing learning rate from $5\times 10^{-3}$ to $5\times 10^{-6}$ and $(\beta_1,\beta_2) = (0, 0.999)$. The annealing schedule we designed for this task reflects the needs for fewer overall steps and is shown in \ref{fig:annealing_schedules} (b). We also decrease the weights of both conditional components of the loss, from $\lambda_T = 1$ to $\lambda_1 = 0$ as we perform the optimization steps $T,\dots,1$ to reduce their influence on the final inferred sample. In addition, we linearly decrease the $\sigma$ parameter that is used to convert the one-hot representations learned from the DDPM model to probabilities, from $\sigma_T=0.2$ to $\sigma_1=0.02$ to mimic the uncertainty of this conversion process. Further examples of land cover segmentation inference are shown in Fig.~\ref{fig:seg_inference_samples_appendix}. Despite the fact that the DDPM was trained only on PA land cover labels we show how the weak label guidance allows us to perform inference in completely new geographies, such as that of AZ (last two rows), where the most prominent label is now Soil and Barren. We can still observe a few artifacts of the PA-related biases the model has learned, like the tendency to add uninterrupted forested areas but the transferability of the semantic model is still far superior than an of an image-based one.

Our hand-crafted constraint for land cover segmentation inference is split between two objectives; (i) matching the structure of the target image using a local color clustering $\vect{z}$ and (ii) forcing the predicted segments’ distribution to match the weak label distribution $\ell_{weak}$ when averaged in non-overlapping blocks of the image.

The local color clustering $\vect{z}$ is computed as a local Gaussian mixture with a fixed number of components. To match the structure between the predicted labels and the precomputed clustering we compute the mutual information between the two distributions in overlapping patches of $31\times31$ pixels. This choice of constraint pushes the inferred land covers segments in a way that they should match locally the color clustering segments. Although this allows us to infer the labels of large structures like roads and buildings it also tends to add noisy labels at areas where the clustering has a high entropy. By gradually reducing the weight of the auxiliary objective however, we allow the inference procedure to `fill in' these details as it is dictated by the diffusion prior.

\begin{figure}[t]
    \centering
    \small
    \begin{tabular}{ccc}
         & Weak &  \\
        Image & Labels $\ell_{\rm weak}$ & Ground Truth \\
        \includegraphics[width=0.15\textwidth]{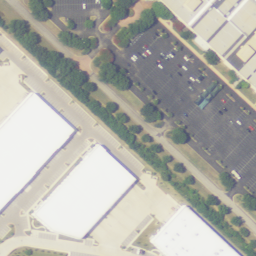} &
        \includegraphics[width=0.15\textwidth]{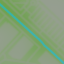} &
        \includegraphics[width=0.15\textwidth]{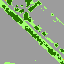} \\
        \\
         & Inferred $\vect{x}$ & Inferred $\vect{x}$ \\
        Clustering $\vect{z}$ & (no guidance) & (with guidance) \\
        \includegraphics[width=0.15\textwidth]{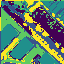} &
        \includegraphics[width=0.15\textwidth]{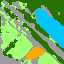} &
        \includegraphics[width=0.15\textwidth]{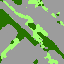} 
    \end{tabular}
    \caption{Inference with and without weak label guidance.}
    \label{fig:label_guidance}
\end{figure}

The label guidance during inference is provided from probabilistic weak labels $\ell_{weak}$ which are derived from coarse auxiliary data. These data are composed of the 30m-resolution National Land Cover Database (NLCD) labels, augmented with building footprints, road networks and waterways/waterbodies \cite{rolf2022resolving}. The corresponding weak label constraint is computed as the KL-divergence between the average predicted and weak label distributions in non-overlapping blocks of $31\times31$ pixels. In the absence of such guidance the inference procedure can easily confuse semantic classes while still producing segmentations that are likely under $p_{DDPM}(x)$. We showcase this in Fig.~\ref{fig:label_guidance} where we infer the land cover labels of an image, starting from a random initialization, with and without the weak label guidance.

\paragraph{Domain transfer.} Regarding the domain transfer experiments, we initially pretrained the standard inference U-Net \cite{unet} on $2\times10^{4}$ batches of 16 randomly sampled $64\times64$ image patches in Pittsburgh, PA, using the Adam optimizer with a learning rate of $10^{-4}$. We then inferred the land cover segmentations of 640 randomly-sampled patches in each of the other geographic regions, (NC, TX, AZ) using the inference procedure described above. With these generated labels, we first finetuned the original network on a validation set of 5 tiles to determine the optimal finetuning parameters. For Durham, NC and Austin, TX we only finetune the last layer of the network for a single epoch, using a batch size of 16 patches and a learning rate of $5\times10^{-4}$. For Phoenix, AZ we require 5 epochs of finetuning the entire network with a learning rate of $5\times10^{-4}$ since the domain shift is larger. Additionally, for all regions, following the experiments of \cite{rolf2022resolving}, we multiply the predicted probabilities with the weak labels and renormalize. 

Finally, in Table 1, we also present the results when the inference network is trained from scratch, to show that the resulting performance is not only an artifact of the pretraining. The U-Net was trained for 20 epochs on all 640 generated samples, with a batch size of 16 and a learning rate of $10^{-3}$.

\subsection{TSP}
\label{sec:tsp_appendix}
\paragraph{DDPM training.} The DDPM was trained on $64\times64$ images of ground truth TSP solutions encoded as images. The architecture was the same U-Net as used in the other experiments, with the architecture from \cite{dhariwal2021diffusion} and $T=1000$ diffusion steps in training. We trained each model for 8 epochs with batch size 16, which took about two days on one Tesla K80 GPU.

\paragraph{Performing inference.} At inference time, we performed varying numbers of inference steps (see Table 2 in the main text), using the Adam optimizer with $(\beta_1,\beta_2)=(0, 0.9)$ and a learning rate linearly decaying from 1 to 0.1. The noise schedule was the same as that used in the MNIST experiment (Fig.~\ref{fig:annealing_schedules}), with the time interval from 0 to 1000 linearly resampled to the number of inference steps used.

To extract a tour from the inferred adjacency matrix $A$, we used the following greedy edge insertion procedure.
\begin{itemize}[topsep=0pt,itemsep=0pt,leftmargin=*]
\item Initialize extracted tour with an empty graph with $N$ vertices. 
\item Sort all the possible edges $(i,j)$ in decreasing order of $A_{ij}/\|v_i-v_j\|$ (i.e., the inverse edge weight, multiplied by inferred likelihood). Call the resulting edge list $(i_1,j_1),(i_2,j_2),\dots$.
\item For each edge $(i,j)$ in the list:
\begin{itemize}[topsep=0pt,itemsep=0pt]
    \item If inserting $(i,j)$ into the graph results in a complete tour, insert $(i,j)$ and terminate.
    \item If inserting $(i,j)$ results in a graph with cycles (of length $<N$), continue.
    \item Otherwise, insert $(i,j)$ into the tour.
\end{itemize}
\end{itemize}
It is easy to see that this algorithm terminates before the entire edge list has been traversed. The tour is refined by a na\"ive implementation of 2-opt, in which, on each step, all pairs of edges in the tour $((i,j),(k,l))$ are enumerated and a 2-opt move is performed if the edges cross. For the `2-opt' baseline, the same procedure is performed using a uniform adjacency matrix.

\paragraph{Results on larger problems.} Extending the results in Table 2 of the main text, we evaluate the model trained on TSP instances with 20 to 50 nodes on problems with 200 nodes. We find an optimality gap of 3.77\% (average number of uncrossing moves 219), compared to 3.81\% for 2-opt (average number of uncrossing moves 115), suggesting that the generalization potential is near-saturated at this problem size. As shown in Fig.~\ref{fig:crowding}, the vertices fill the $64\times64$ image with such high density that it is difficult to see the (light grey) tour; many edges are invisible (compare to Fig.~7 in the main text).

\begin{figure}[t]
    \centering
    \begin{tabular}{@{}c@{}c@{}c@{}c@{}c@{}c@{}c@{\hskip1pt}c@{\hskip1pt}c@{}}
        &\multicolumn{5}{c}{\small Optimize latent adjacency\ matrix w.r.t.\ denoising model}
        &\multicolumn{2}{c}{\small Recover tour}
        \\\cmidrule(lr){2-6}\cmidrule(lr){7-8}
            \small Input& \small $t=$256 & \small $t=$192& \small $t=$128 & \small $t=$64&\small $t=$0 & \small Extracted & + 2-opt &  \small Oracle\\
        \includegraphics[width=0.11\textwidth,trim=0 0 0 0,clip]{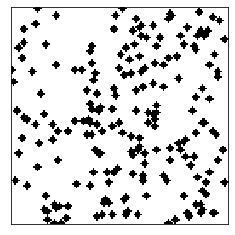}&
        \includegraphics[width=0.11\textwidth,trim=0 0 0 0,clip]{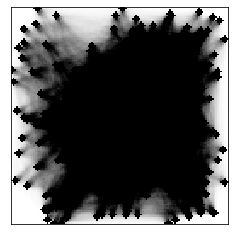}&
        \includegraphics[width=0.11\textwidth,trim=0 0 0 0,clip]{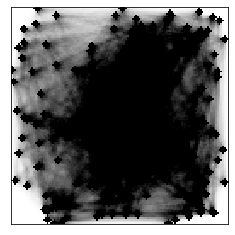}&
        \includegraphics[width=0.11\textwidth,trim=0 0 0 0,clip]{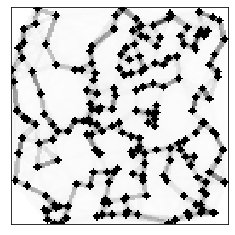}&
        \includegraphics[width=0.11\textwidth,trim=0 0 0 0,clip]{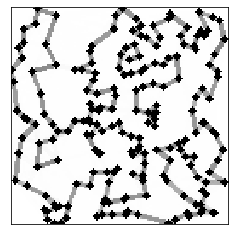}&
        \includegraphics[width=0.11\textwidth,trim=0 0 0 0,clip]{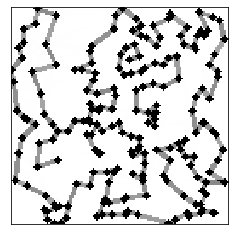}&
        \includegraphics[width=0.11\textwidth,height=0.11\textwidth,trim=5 0 5 0,clip]{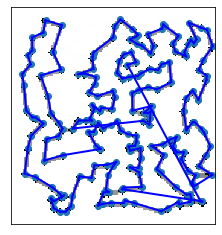}&
        \includegraphics[width=0.11\textwidth,height=0.11\textwidth,trim=5 0 5 0,clip]{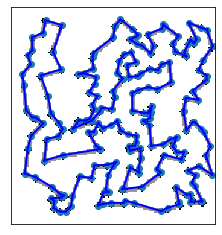}&
        \includegraphics[width=0.11\textwidth,height=0.11\textwidth,trim=5 0 5 0,clip]{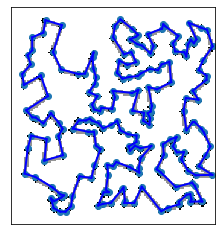}\\
    \end{tabular}
    \caption{Latent adjacency matrix inference in a 200-vertex TSP, using a model trained on $64\times64$ images but $128\times128$ images at inference time. The discovered tour is 2.12\% longer than the optimal one.}
    \label{fig:tsp_latent_big}
\end{figure}

We suggest three directions to solving to this problem that should be explored in later work:
\begin{enumerate}[left=0pt,label=(\arabic*)]
    \item Encoding: The size of the encoding image can be increased (for example, to $128\times128$) when the number of vertices increases, without changing the model (trained on $64\times64$ images), which can make denoising predictions on images of any size. We may expect to see better out-of-domain generalization of the denoising model in this setting, as the density of nodes (mean number of black pixels) would match that in the training set. Figs.~\ref{fig:tsp_gen_big} and \ref{fig:tsp_latent_big} show the potential of DDPMs to generalize to image sizes larger than those in which they were trained. Inference using $128\times128$ images gives an optimality gap of 2.59\% (average number of uncrossings 81), much lower than that obtained with in-domain image size.
    
    In addition, encoding graphs with smaller dots and thinner lines can be explored, although the generalization difficulties due to image `crowding' would still appear at a larger value of $N$.
    \item Fractal behaviour and coarse-to-fine: Taking advantage of the fractal structure of Euclidean TSP solutions, a denoising objective could be used to \emph{locally} refine the tour by minimizing the objective on a \emph{crop} of the image representation (a form of DDPM-guided local search). This could be done in a coarse-to-fine manner by application of the same model at different scales, with a $128\times128$ representation of a problem with 200 vertices being first optimized with respect to the denoising objective globally, then on $64\times64$ crops.
    \item Improved extraction: The 2-opt search can be improved by inexpensive heuristics, such as choosing the 2-opt move that most improves the cost on every step, rather than iterating through the edges of the candidate tour in order.
\end{enumerate}

\begin{figure}[b]
    \centering
    \begin{minipage}[b]{0.5\textwidth}
        \centering
        \begin{tabular}{cc}
            \small Input & \small Solution \\
            \includegraphics[width=0.4\textwidth]{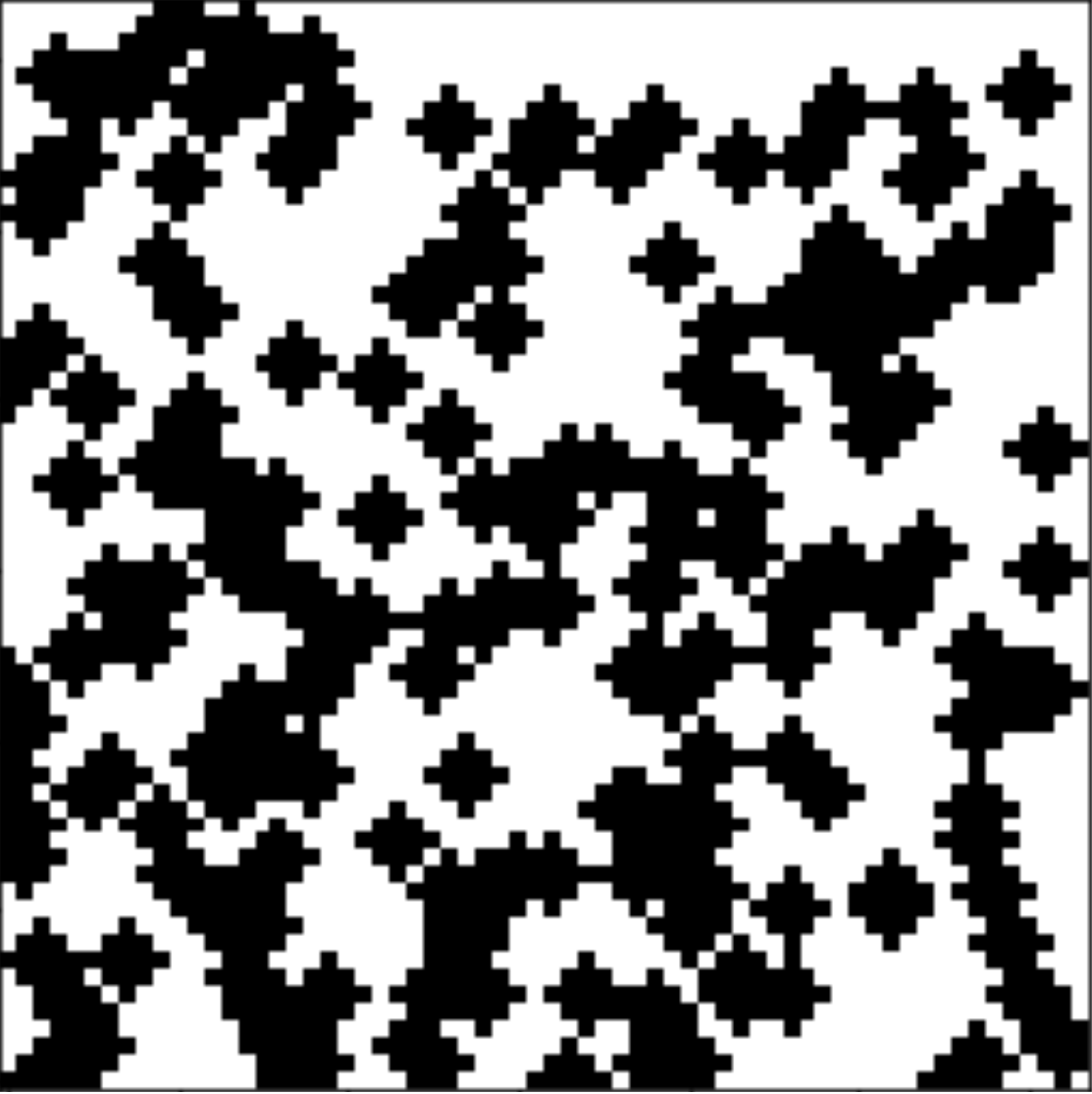} &
            \includegraphics[width=0.4\textwidth]{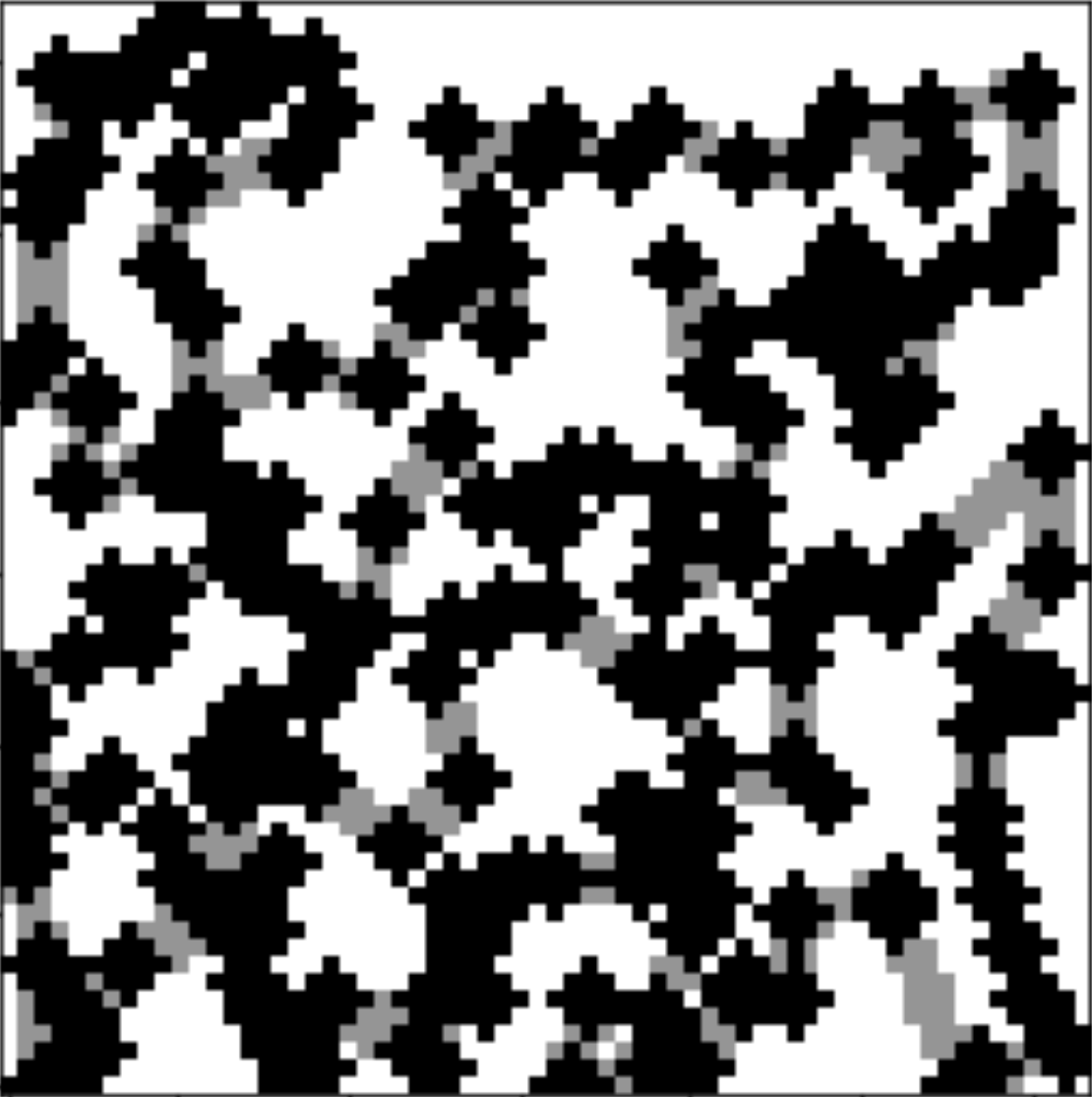} \\
        \end{tabular}
        \caption{A TSP instance and the ground truth solution with $N=200$ vertices encoded in a $64\times64$ image.}
        \label{fig:crowding}
    \end{minipage}
    \hfill
    \begin{minipage}[b]{0.45\textwidth}
        \centering
        \vfill
        \includegraphics[width=0.5\textwidth]{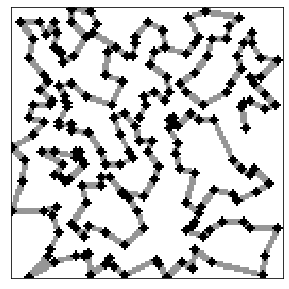}
        \caption{Unconditional $128\times128$ samples from the DDPM trained on $64\times64$ image representations of 50-vertex TSPs.}
        \label{fig:tsp_gen_big}
    \end{minipage}
\end{figure}

\end{document}